\documentclass[journal]{IEEEtran}

\usepackage[utf8]{inputenc}
\usepackage{booktabs,tabularx}
\usepackage{times}
\usepackage{epsfig}
\usepackage{graphicx}
\usepackage{amsmath}
\usepackage{amssymb}
\usepackage{pifont}
\usepackage[table,usenames,dvipsnames,svgnames]{xcolor} 
\usepackage{colortbl}
\usepackage{wrapfig}
\usepackage{tabu}
\usepackage{algorithm}
\usepackage[noend]{algpseudocode}
\usepackage{upgreek}
\usepackage{cite}
\graphicspath{{images/}}

\makeatother

\usepackage{blindtext}

\newcommand{\dashrule}[1][black]{%
	\color{#1}\rule[\dimexpr.5ex-.2pt]{4pt}{.4pt}\xleaders\hbox{\rule{4pt}{0pt}\rule[\dimexpr.5ex-.2pt]{4pt}{.4pt}}\hfill\kern0pt%
}
\newcommand{\rulecolor}[1]{%
	\def\CT@arc@{\color{#1}}%
}

\usepackage{booktabs}
\usepackage{multirow}

\usepackage{stackengine}
\usepackage{scalerel}
\newlength\lthk
\definecolor{verylightgray}{gray}{0.95}
\newlength{\tabwidth}

\newcommand{\proc}[1]{\textcolor[rgb]{0.64,0,0}{#1}}
\newcommand{\abso}[1]{\textcolor[rgb]{0,0.6875,0.93}{#1}}

\usepackage[unicode,hyperindex,plainpages=false,pdftex,hidelinks]{hyperref}
\hypersetup{
	colorlinks=true,
	linkcolor=BrickRed,
	citecolor=OliveGreen,
	filecolor=magenta,
	urlcolor=cyan
}

\newcommand{\CarsTotalCount}{20\,865 }

\begin{document}
%
\title{Comprehensive Dataset for Automatic Single Camera Visual Speed Measurement}
%
%
%

\author{Jakub Sochor, Roman Juránek, Jakub Špaňhel, Lukáš Maršík, Adam Široký, Adam Herout, Pavel Zemčík
\thanks{Authors are with Brno University of Technology, Faculty of Information Technology, Centre of Excellence IT4Innovations , Czech Republic \{isochor,ijuranek,herout\}@fit.vutbr.cz}
\thanks{Jakub Sochor is a Brno Ph.D. Talent Scholarship Holder --- Funded by the Brno City Municipality.}
}

\markboth{IEEE Transactions on Intelligent Transportation Systems}%
{Sochor \MakeLowercase{\textit{et al.}}: Comprehensive Dataset for Automatic Single Camera Visual Speed Measurement}


\maketitle

\begin{abstract}
In this paper, we focus on traffic camera calibration and visual speed measurement from a single monocular camera, which is an important task of visual traffic surveillance.
Existing methods addressing this problem are hard to compare due to a lack of a common dataset with reliable ground truth. Therefore, it is not clear how the methods compare in various aspects and what are the factors affecting their performance.
We captured a new dataset of 18 full-HD videos, each around one hour long, captured at 6 different locations. Vehicles in the videos (\CarsTotalCount instances in total) are annotated with precise speed measurements from optical gates using LIDAR  and verified with several reference GPS tracks. We made the dataset available for download and it contains the videos and metadata (calibration, lengths of features in image, annotations, etc.) for future comparison and evaluation.
Camera calibration is the most crucial part of the speed measurement; therefore, we provide a brief overview of the methods and analyze a recently published method for fully automatic camera calibration and vehicle speed measurement and report the results on this dataset in detail.
%
\end{abstract}

\section{Introduction}
Speed measurement is one of the crucial problems in traffic surveillance.
So far, the field is dominated by radar and section speed measurements because they meet tight methodological requirements and standards.
However, these methods are li\-mi\-ted in the information they provide and they may be expensive.
For example, in radar measurement it is impossible to recognize the fine-grained models of passing cars and the radar antenna must be placed at a specific position regarding the traffic.
Section speed measurement requires two cameras for each position and a complex infrastructure for processing the data.
Speed measurement from a single monocular camera is not typically used for surveillance; however it can be beneficial -- one camera can be used for surveillance on multiple lanes, it is possible to use the data for fine-grained make\,\&\,model recognition of the vehicles \cite{He2015,Baran2015,Hu2015ITS,Hsieh2014} and other tasks. Another interesting aspect is that it is possible to use already installed monitoring/security cameras for speed measurement and other traffic analysis tasks.


\begin{figure}[t]
	\centering
	\def\svgwidth{0.8\linewidth}
	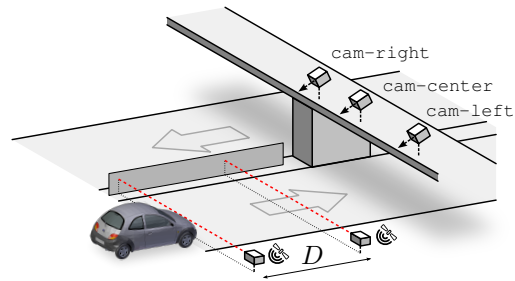
	\caption{
		The data recording setup. We use two LIDARs	synced by GPS time, and three cameras recording the highway from different surveillance viewpoints.
	}
	\label{fig:Setup}
\end{figure}

A number of works dealing with monocular speed measurement can be found in the literature \cite{Schoepflin2003,Dailey2000,Grammatikopoulos2005,He2007,Maduro2008,Nurhadiyatna2013,Sina2013,Dubska2014,Lan2014,Luvizon2014,Do2015} (detailed individually below).  Such systems are on the rise especially recently, with the growing number of IP cameras, with increase of their resolution, and with the development of computer vision algorithms used for their processing.  Our aim is to provide an important missing piece: a dataset which would allow for reliable comparison between the approaches. These systems are described in detail in the following section.

We captured a new benchmark dataset of 18 full-HD videos taken from surveillance viewpoints on the traffic (see Figure~\ref{fig:Setup}).
Each of the videos is around one hour long to allow for even lengthy calibration procedures and self-adjustment of the surveillance system.  Triplets of videos are observing the same time interval at the same location from different angles.  These shots were captured at 6 different locations. Vehicles in the videos (\CarsTotalCount instances in total) are annotated with precise speed measurements from optical gates using LIDAR and verified with several reference GPS tracks. We provide%
\footnote{\url{https://medusa.fit.vutbr.cz/traffic}}
the videos and metadata (calibration, distances measured on the road plane, annotations, etc.) for future comparison and evaluation.
To illustrate the properties of the dataset and to establish a first baseline, we analyze the data by a recently published method for fully automatic camera calibration and vehicle speed measurement \cite{Dubska2014} and we report the quantitative results.


Although the dataset is focused on speed measurement, it can be used also for different traffic surveillance tasks, for example vehicle counting, tracking, vehicle classification and other.



We consider the camera calibration algorithm to be the most crucial part of speed measurement. It defines how well the speed measurement is done as it is impossible to measure speed accurately with a poorly calibrated camera. The used algorithm also defines whether it is usable with a camera observing the road from arbitrary viewpoint and it determines whether the method can be used fully automatically which is important for large scale deployment. Therefore, we include a~brief overview of existing camera calibration algorithms for traffic surveillance applications.

The key contributions of this paper are:
\textbf{a)} Novel, publicly available dataset for evaluation of camera calibration in traffic surveillance and speed measurement. The dataset contains 18 videos and \CarsTotalCount vehicles with known precise ground truth.
\textbf{b)} Thorough and complex evaluation of a recent fully automatic method for traffic camera calibration \cite{Dubska2014}.

\section{Related Work -- Camera Calibration for Speed Measurement of Vehicles}
\label{sec:SOTASpeedMeasurement}

\begin{figure}[t]
	\centering
	\def\svgwidth{\linewidth}
	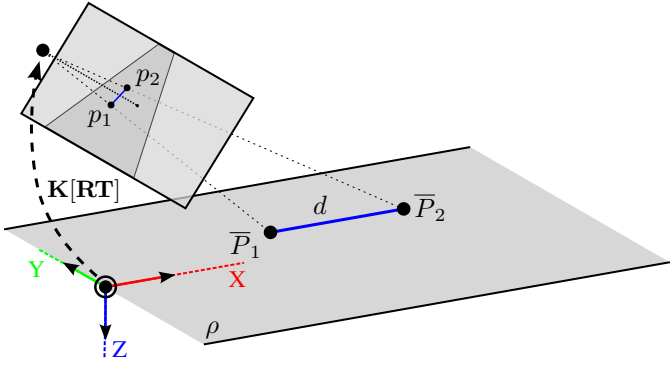
	\caption{
		The essential goal of traffic surveillance camera calibration is to be able measure real world distance $d$ between two points ($\overline{P}_1$, $\overline{P}_2$) on road plane given their projection to the image ($p_1$, $p_2$). X,Y, and Z axes represent a real world coordinate system and $\mathbf{K}$ represent intrinsic camera parameters, while $\mathbf{R}$ and $\mathbf{T}$ are extrinsic camera parameters.
	}
	\label{fig:Projection}
\end{figure}

One of the most important parts of speed measurement of vehicles from a single monocular camera is calibration of the camera. In a general case, this includes dealing with perspective projection and different rotations of the camera; it is also necessary to deal with unknown distance from the camera to the ground plane of the road and possibly with radial and tangential distortion. It is usually necessary to obtain intrinsic and extrinsic camera parameters together with the scene scale (or the distance of the camera from the road/ground plane). Therefore, we include also a brief overview of the typical solutions of camera calibration for speed measurement of vehicles. However, first we include the definition of traffic surveillance camera calibration. 

\subsection{Traffic Surveillance Camera Calibration}
General mathematical model for camera calibration is represented by a projection matrix $\mathbf{P} = \mathbf{K~[R~T]}$, where $\mathbf{K}$ denotes intrinsic camera parameters,  $\mathbf{R}$ stands for camera rotation, and $\mathbf{T}$ represent camera translation. The extrinsic parameters (rotation and translation) are relative to defined world coordinate system (see Figure \ref{fig:Projection}). Since the calibration for traffic surveillance is specific, we describe all these aspects with \textbf{application to vehicle speed measurement} in mind.

\paragraph{Goal} The essential goal of traffic surveillance camera calibration is to measure speed of vehicles. For the speed measurement, it is required to be able measure time and distances on the road plane. The time measurement part is rather trivial. However, for the distance measurement it is necessary to measure the distance between two points on the road plane (or any other plane parallel to the road plane and with known distance from the road plane) given their projection to the image. See Figure \ref{fig:Projection} for an example. 

\paragraph{Input} For fully automatic methods, the camera calibration input is usually a video of the observed traffic scene. However, for methods which include manual steps, part of the input are also usually distance measurements on the road plane. 

\paragraph{Assumptions} Zero pixel skew is generally used as an assumption about the camera model. Another widely used assumption is that the camera's principal point is in the center of the image. Also, there is usually the assumption that the road can be approximated by a plane. The authors usually assume that the observed road segment is approximately straight, that the vehicles move straight and their velocity is constant on the measured segment (no acceleration).

\paragraph{Mathematical model}
As the standard camera model with $\mathbf{K~[R~T]}$ matrices is sufficiently described in existing literature \cite{Hartley2004}; we refer the readers there. However, it is also possible to use a different formulation based on vanishing points of the road plane \cite{Dubska2015ITS}. This formulation is easily convertible to the standard one. Finally, for computation of 3D real world coordinates on the road plane of a point in image space, it is necessary to compute intersection of the road plane (e.g. $z=0$ as shown in Figure \ref{fig:Projection}) and a ray defined by the camera optical center and the coordinates on the image plane. 

\paragraph{Attributes}
One important attribute of the camera calibration algorithm, which should be kept in mind, is whether the algorithm works automatically in the sense that there is no manual input required per installed camera. The property of being automatic becomes more important as the number of installed cameras grows. A number of papers and approaches to solving this problem exist and they will be discussed in detail in the following text. Another important attribute is whether the algorithm works from arbitrary viewpoint, as it is a significant drawback of a method if it requires specific camera placement relative to the observed road.

\subsection{Methods Based on Acquired Line Markings}
He and Yung \cite{He2007} proposed a method for speed estimation of vehicles which is based on calibration using a calibration pattern formed by lane markings on the road \cite{He2007CalibMethod}. The authors use a rectified image in further processing in order to deal with perspective projection. To obtain the locations of the vehicles within the ground plane, shadows cast by rear bumpers are used. The vehicles and shadows are detected by background subtraction and binary block matching.

Cathey and Dailey \cite{Cathey2005} used a method based on detection of the vanishing point which is in the direction of vehicles movement. To obtain this vanishing point, detected line markings are used and their intersection is found in the least squares manner. The scale (pixels/meters ratio) for the camera is computed from average line marking stripe length and known stripe length in the real world. Finally, the authors used cross correlation to compute the number of pixels which vehicles passed between consecutive frames.

Grammatikopoulos et al.\,\cite{Grammatikopoulos2005} use the assumption that the camera is only tilted along Y axis in Figure \ref{fig:Projection}; thus they assume that the \emph{second vanishing point} (horizontal and perpendicular to the first one) is in infinity. The first vanishing point is detected as the intersection of the line markings with least squares adjustment. The vehicles are detected by background subtraction and tracked by normalized cross-correlation.

You et al.~\cite{You2016} propose to use detection of vanishing point in the direction of vehicles' movements from lane markings and vanishing point perpendicular to road plane from detected poles and pedestrians. The authors obtain the scale from known height of the camera above the road or known dimensions on the road.

By definition, this class of methods based on observed line markings is usable only when the line markings are present, visible, and recognizable. This fact can be limiting on local roads, where the line markings are not drawn or on highways during work on the road with additional temporary line markings. Also, some of the methods require measurements on the road which is a great disadvantage.

\subsection{Methods Based on Vehicles' Movement}
Dubská et al.\,\cite{Dubska2014} published a speed measurement system using a calibration method by detection of two vanishing points \cite{Dubska2015ITS}. We give the details on this method below in Section \ref{sec:method}.

Schoepflin et al.\,\cite{Schoepflin2003} use an activity map (by detecting the vehicles as the moving foreground) to obtain lane boundaries and the intersection of the boundaries treat as the first vanishing point in the direction of the vehicle motion. The second vanishing point is detected as the intersection of lines formed by the bottom edges of the vehicles. One known length (manually measured and entered per camera) in the image is used for scale inference.

Filipiak et al.~\cite{Filipiak2016} propose to use sequences of detected license plates of vehicles for finding intrinsic and extrinsic camera parameters by an evolutionary algorithm. The method was evaluated on a dataset captured by zoomed surveillance cameras with a small field of view on the road.

The methods based on vehicles' movement no longer need visible road markings; however, when used on small local roads, the calibration may take some time as it usually improves with more observed vehicles. 

\subsection{Methods Using Manual Measurements}
Maduro et al.\,\cite{Maduro2008} assume two known arbitrary angles on the ground plane to calibrate the camera and use lengths of line markings' stripes to obtain the camera scale for the given scene. The authors used background subtraction for detecting the vehicles and Kalman filter \cite{Kalman1960} for tracking them.

Nurhadiyatna et al.\,\cite{Nurhadiyatna2013} used GMM background subtraction~\cite{Zivkovic2004} for detection of vehicles and tracked them by Kalman filter. They use a calibrated pinhole camera with zero pan and known distances in the real world.

Sina et al.\,\cite{Sina2013} focus on speed measurement at night. They used detected and paired headlights to detect vehicles, track them and measure their speed. The camera calibration is based on manual measurements of camera angles and distance of the camera from the ground plane. The reported average error is 3.3\,km/h relative to ground truth obtained by GPS.

Luvizon et al.\,\cite{Luvizon2014} used a different approach and they propose to detect and track license plates in order to obtain motion of vehicles in the scene. The motion is then converted to the real world distance by rectifying and scaling. The scale inference is based on a priori known real world measures.

Methods using manual measurements on the road have the biggest disadvantage that it is necessary to do the manual measurements, which potentially can mean stopping traffic on the road. The advantage of the methods may be (in some cases) that they are more accurate than automatic or semi-automatic ones.

\begin{table*}[!tpb]
	\caption{Summary of different camera calibration methods for speed measurement. It should be noted that the reported errors are only informative as all the methods are evaluated on different datasets and by different protocols. We consider a system to be automatic if it does not require any manual calibration for each individual camera. \textbf{auto} -- denotes whether the system works fully automatically, \textbf{view} -- denotes whether the system is usable from arbitrary viewpoint} \label{tab:CamCalibComparison}
	\def\tabularxcolumn#1{m{#1}}
	\begin{center}
		\rowcolors{1}{verylightgray}{white}
		\noindent \begin{tabularx}{\linewidth}{ c X c c c}
			\toprule
			\hiderowcolors
			& \textbf{camera calibration method} & \textbf{auto} & \textbf{view} & \textbf{mean error}
			\\\midrule
			\showrowcolors
			Dailey et al., 2000
			\cite{Dailey2000} & multiple assumptions on vehicle movements and known mean length of vehicles & \ding{51} & \ding{55} & 6.5\,km/h\\
			Schoepflin et al., 2003
			\cite{Schoepflin2003} & detection of two vanishing points, one known length & \ding{55} & \ding{51} & N/A\\
			Cathey et al., 2005
			\cite{Cathey2005} & vanishing point obtained from detected line markings, scale computed from lengths of stripes & \ding{55} & \ding{51} & N/A\\
			Grammatik. et al., 2005
			\cite{Grammatikopoulos2005} & one vanishing point obtained from detected line markings, the second one assumed in infinity, one known distance is required & \ding{55} & \ding{55} & 3\,km/h\\
			He and Yung, 2007
			\cite{He2007} & calibration by pattern formed by lane markings & \ding{55} & \ding{51} & 3.27\,\%\\
			Maduro et al., 2008
			\cite{Maduro2008} & known angle of the ground plane, lengths of line markings' stripes & \ding{55} & \ding{51} & 2\,\%\\
			Nurhadiyatna et al., 2013
			\cite{Nurhadiyatna2013} & known distances in the real world and in the scene, zero pan assumption & \ding{55} & \ding{55} & 7.63\,km/h\\
			Sina et al., 2013
			\cite{Sina2013} & manual measurements & \ding{55} & \ding{51} & 3.3\,km/h\\
			Dubská et al., 2014
			\cite{Dubska2014}  &  detection of two vanishing points, scale computed by matching of statistics of vehicles' dimensions to mean dimensions of vehicles & \ding{51} & \ding{51} & 1.99\,\%\\
			Lan et al., 2014
			\cite{Lan2014} & relaxation of perspective projection, known width of lanes & \ding{55} & \ding{55} & 0.9\,\% -- 2.5\,\%\\
			Luvizon et al., 2014
			\cite{Luvizon2014} & known real world measures & \ding{55} & \ding{51} & 1.63\,km/h\\
			Do et al., 2015
			\cite{Do2015} & zero pan assumption, equilateral triangle drawn on the road & \ding{55} & \ding{55} & 2.91\,\%\\
			Filipiak et al., 2016
			\cite{Filipiak2016} & constant speed assumption, evolutionary algorithm to recover intrinsic and extrinsic parameters from detected license plate sequences & \ding{51} & \ding{55} & 2.3\,km/h \\
			You et al., 2016
			\cite{You2016} & detection of vanishing point in the direction of vehicles' movements from lane markings and vanishing point perpendicular to road plane from poles and pedestrians, the scale is obtained from known height of camera above the road & \ding{55} & \ding{51} & N/A \\
			\bottomrule
		\end{tabularx}
	\end{center}
	
\end{table*}

\subsection{Automatic Calibration Method based On Statistics of Dimensions}
\label{sec:method}
Here we give details on the speed measurement method of Dubská \cite{Dubska2014}, as it meets all our requirements (it is fully automatic and it is usable from arbitrary viewpoint) and we use it later in the experiments.
In principle, the method relies on camera calibration from two automatically detected vanishing points.

\begin{figure}[h]
	\centering
	\includegraphics[width=0.3\linewidth]{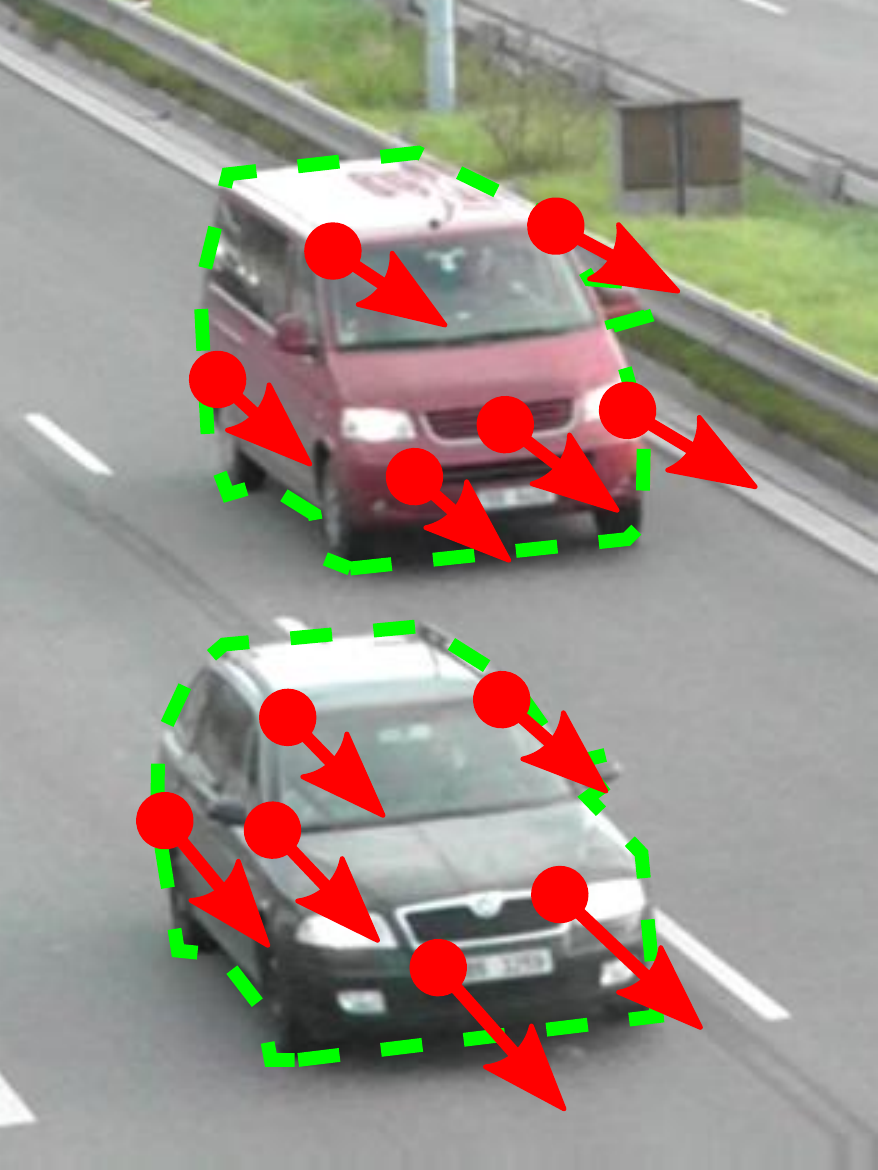}\hfill
	\includegraphics[width=0.3\linewidth]{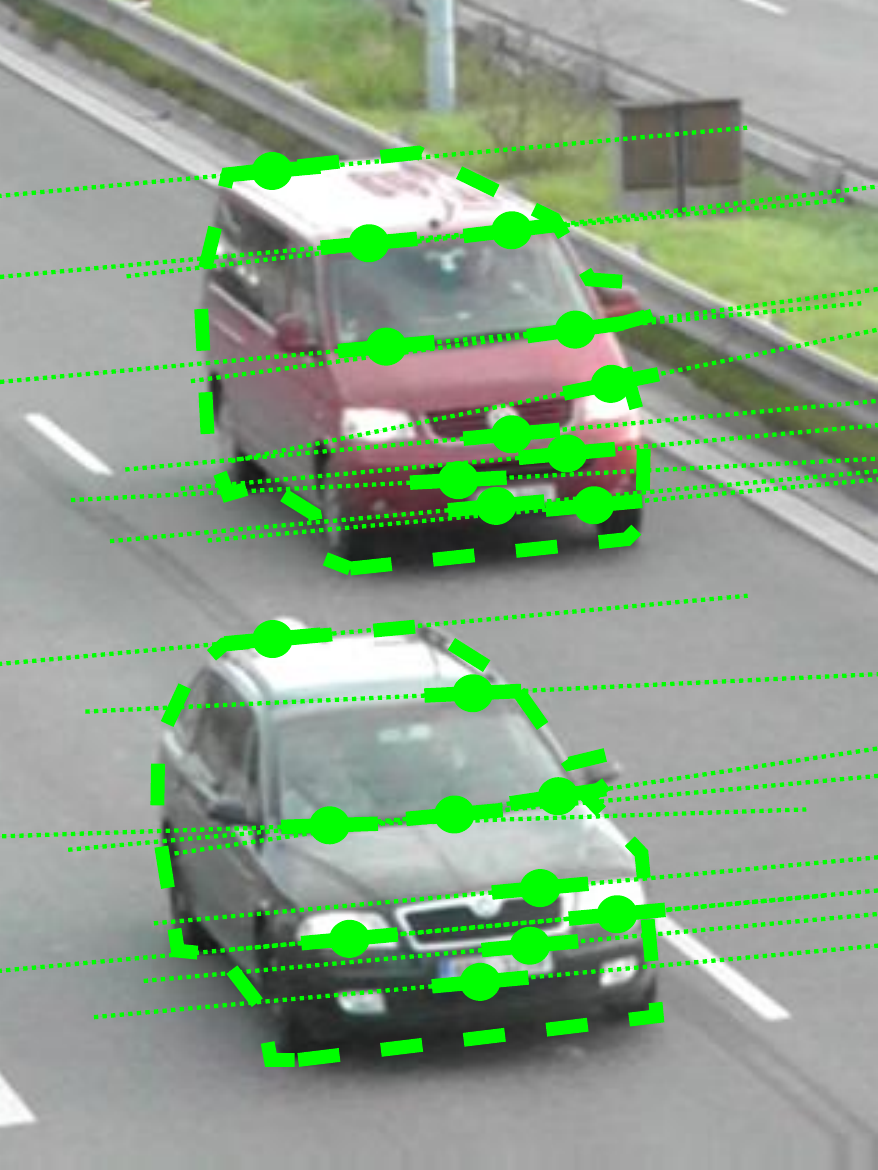}\hfill
	\includegraphics[width=0.3\linewidth]{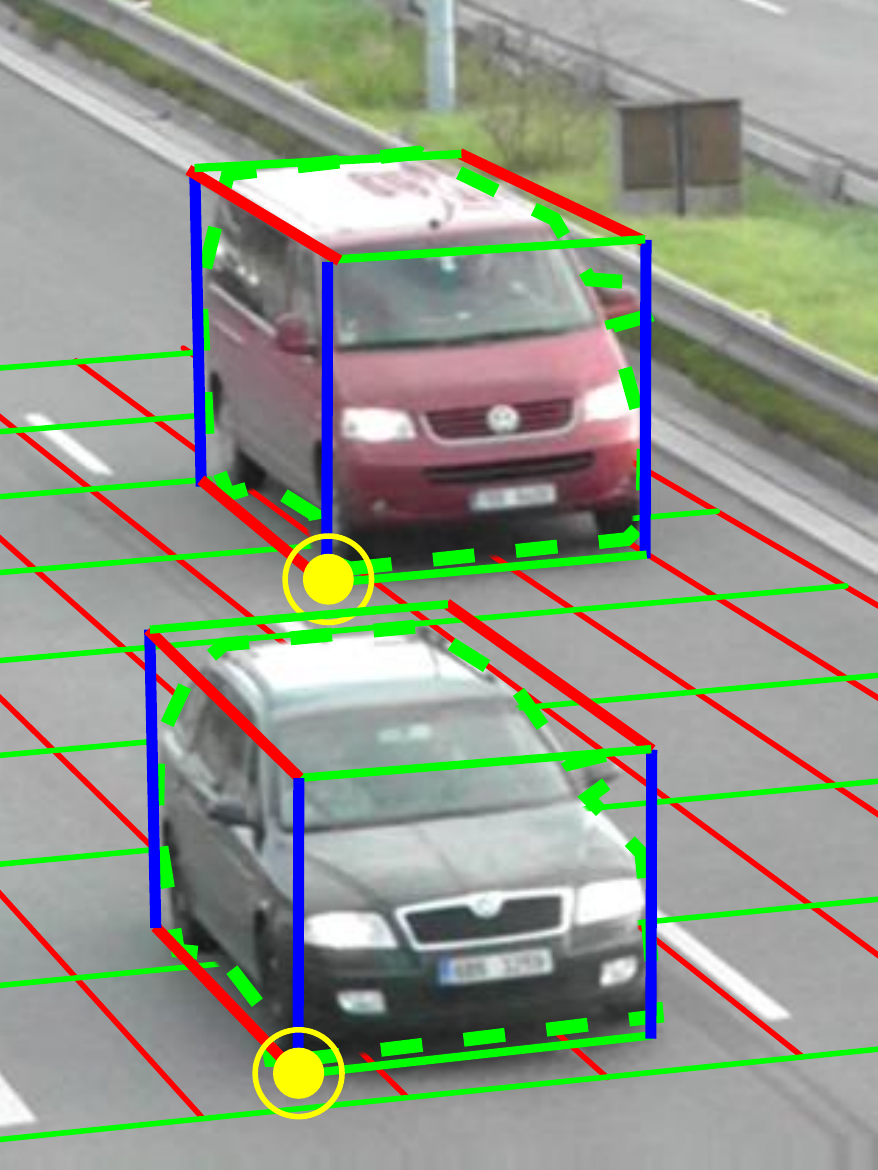}
	\caption{Automatic camera calibration according to Dubska~\cite{Dubska2014}. From left to right: Tracked keypoints for VP1, oriented edges voting for VP2, and road plane with bounding boxes for the cars and reference points for tracking.}
	\label{fig:MethodScheme}
\end{figure}

The authors use a simple foreground detection model to filter areas with movement.
The first vanishing point (VP1, which is in the direction of vehicles' movement) is re\-co\-ve\-red from tracked feature points on the vehicles using min eigenvalue detector and KLT tracker. The tracked points' motion is transformed using a  line-to-line Hough transformation parametrized by parallel coordinates \cite{Dubska2013}  where the global maximum corresponds to the image of the first vanishing point.
The second vanishing point (VP2) is extracted from strong edges present on the moving vehicles meeting some conditions given by the position of the VP1.  The edges (and their orientations) are, again, transformed to the Hough space where the strongest maximum accounts for the vanishing point.
From these two vanishing points, the camera intrinsics and extrinsics can be recovered (assuming principal point in the image center, square pixels and zero skew).

The authors propose an algorithm for computing the 3D bounding box around the vehicle blobs. Mean size of the bounding boxes and known mean dimensions of the vehicles for a given country accounts for the scene scale.
Vehicle speed is measured simply by tracking 3D bounding boxes around the blobs using Kalman filter and measuring the travel distance in the real world.

The authors evaluated their method on several videos with several car passes with ground truth speed obtained from GPS.

\subsection{Other Methods}
Dailey et al.~\cite{Dailey2000} proposed a method for vehicles speed measurement based on tracking of vehicle blobs and constraining them to move along a line. The blobs are detected as inter-frame differences followed by Sobel edge detector. The authors assume that the vehicles are moving towards or from the camera and use mean length of vehicles to obtain the scene scale.

Do et al.~\cite{Do2015} proposed a camera calibration method for speed measurement based on artificial markers drawn on the road. They assume that the camera has zero pan angle and that markers determining vertices of an equilateral triangle with a known distance between vertices which are visible on the road. They used the triangle to obtain the scale factor and the tilt angle.

Lan et al.~\cite{Lan2014} use optical flow to compute the speed of different points of a vehicle and they average this speed to get the speed of vehicle in image units. However, to convert them into kilometers per hour, the authors assume that there is no perspective projection effect and the width of the ROI (width of lanes) is known.

\begin{table*}[!t]
	\centering
	\caption{Summary of datasets used for evaluation of visual speed measurement methods.} \label{tab:SOTADatasets}
	\def\tabularxcolumn#1{m{#1}}
	\rowcolors{1}{verylightgray}{white}
	\noindent \begin{tabularx}{\linewidth}{ c r r c c X }
		\toprule
		\hiderowcolors
		\textbf{dataset} & \textbf{videos} & \textbf{vehicles} & \textbf{source of gt} & \textbf{resolution} & \textbf{evaluation metrics}\\
		\midrule
		\showrowcolors
		Dailey et al., 2000
		\cite{Dailey2000} & 1 & 532 & induction loops & N/A & speed measurement error \\
		Schoepflin et al., 2003
		\cite{Schoepflin2003} & 2 & 1\,015 & induction loops & $320 \times 240$ &  speed measurement error  \\
		Grammatik. et al., 2005
		\cite{Grammatikopoulos2005} & 1 & 20 & manual measurements  & $768 \times 576$ &  speed measurement error\\
		He and Yung, 2007 
		\cite{He2007} & 1 & 64 & RADAR & $1280 \times 1024$ & speed measurement error\\
		Maduro et al., 2008
		\cite{Maduro2008} & 2 & few & GPS & N/A & speed measurement error\\
		Nurhadiyatna et al., 2013 
		\cite{Nurhadiyatna2013} & 10 & 15 & GPS & $320 \times 240$ & speed measurement error\\
		Sina et al., 2013 
		\cite{Sina2013} & 13 & 13 & GPS & N/A & speed measurement error, vehicle counting\\
		Dubská et al., 2014 
		\cite{Dubska2014} & 6 & 29 & GPS & $864 \times 480$ & speed measurement error, distance measurement error\\
		Lan et al., 2014
		\cite{Lan2014} & 1 & 2\,010 & RADAR & $640 \times 480$ & speed measurement error\\
		Luvizon et al., 2014
		\cite{Luvizon2014} & 1 & 75 & induction loops & $768 \times 480$ & speed measurement error, license plate detection\\
		Do et al., 2015
		\cite{Do2015} & 1 & 3 & speedometer & N/A& speed measurement error\\
		Filipiak et al., 2016
		\cite{Filipiak2016} & 2 & 955 & induction loops & $1280 \times 720$ &  speed measurement error\\
		\midrule
		\textbf{proposed} & \textbf{18} & \textbf{\CarsTotalCount} & \textbf{LIDAR gates} & $\mathbf{1920 \times 1080}$ & \textbf{calibration error, distance measurement error, speed measurement error, vehicle counting recall, false positives vehicles per minute}\\
		\bottomrule
	\end{tabularx}
\end{table*}

\subsection{Summary and Analysis of the Methods}
A summary of the presented camera calibration methods can be found in Table~\ref{tab:CamCalibComparison}. As the table shows, some of the approaches have different limitations and they do not work under all conditions.  The reported mean error varies greatly -- it should be noted that the error is not directly comparable, as it was evaluated by the authors on different datasets (generally not publicly available) and by different protocols.

To sum up the camera calibration methods, some of them \cite{Dailey2000,Lan2014} do no take perspective projection into account, some algorithms \cite{Dailey2000,Grammatikopoulos2005,Nurhadiyatna2013,Lan2014,Do2015} have limitations in camera placement. Quite a large number of approaches \cite{Maduro2008,Nurhadiyatna2013,Sina2013,Luvizon2014} use measurements in the scene which enable direct camera calibration. Methods \cite{He2007,Do2015}  using a calibration pattern (virtual or drawn on the road) have been proposed. Another set of methods use vanishing points to obtain camera calibration \cite{Schoepflin2003,Cathey2005,Grammatikopoulos2005,Dubska2014}.

Several approaches to scale calibration have been proposed. Besides the multiple manual measurements on the road \cite{Maduro2008,Nurhadiyatna2013,Sina2013,Luvizon2014} and calibration patterns \cite{He2007,Do2015}, two groups of methods exist. The algorithms from the first one \cite{Schoepflin2003,Cathey2005,Grammatikopoulos2005,Lan2014} use one known distance in the scene (e.g. length of line marking stripe). The other methods use dimensions of vehicles \cite{Dailey2000,Dubska2014} to obtain a proper scale calibration.

One important attribute of the calibration methods is whether they work fully automatically and do not require any manual per camera calibration input. The automation helps reduce the cost of camera installation and the automatic methods have better scaling properties. Only two approaches are fully automatic and do not require any manual camera calibration. Both of these methods \cite{Dailey2000,Dubska2014} use mean dimension of vehicles to obtain a proper scaling factor for the given camera.

Methods \cite{Maduro2008,Nurhadiyatna2013,Sina2013,Luvizon2014,He2007,Do2015} which require measurements of physical dimensions on the road have even more significant drawbacks with respect to the scaling properties. To perform the measurements, it is usually necessary to stop (or limit) traffic on the road increasing installation time and costs. 

Another important attribute is whether the camera can be placed at any position above the road, as some methods require for example that the camera has zero pan. In real world scenarios, this can be hard to guarantee when the camera is not placed on a portal above the road. The only method that satisfies the conditions of automatic calibration and arbitrary view is \cite{Dubska2014} which we use later in the experiments.


\subsection{Evaluation Datasets Used in Existing Works}
The described methods usually used different methods for evaluation of the speed measurement and ground truth speed acquisition. Some methods \cite{Dailey2000,Schoepflin2003,Luvizon2014,Filipiak2016} use inductive loops for ground truth acquisition, other methods \cite{Maduro2008,Dubska2014} GPS or RADAR \cite{Lan2014}. Do et al. \cite{Do2015} used the speedometer on a motorbike, which should be considered very imprecise.

When it comes to the number of evaluated speed measurements, Lan et al. \cite{Lan2014} used 2\,010 ground truth speeds (only one video sequence), others \cite{Dailey2000,Schoepflin2003,Filipiak2016} have hundreds of vehicles with known ground truth.
And there are also works \cite{Grammatikopoulos2005,He2007,Maduro2008,Nurhadiyatna2013,Sina2013,Dubska2014,Lan2014,Luvizon2014,Do2015} that use at most tens of ground truth speeds with the lowest number in \cite{Do2015} (one ground truth speed) and the highest number of 75 measurements in \cite{Luvizon2014}. Cathey at al. \cite{Cathey2005} have no evaluation at all. A summary of existing datasets can be found in Table~\ref{tab:SOTADatasets}. It should be noted that with the exception of \cite{Nurhadiyatna2013,Dubska2014}, the datasets are not publicly available which makes comparison of the methods impossible.

Almost every mentioned dataset (except \cite{Sina2013} and a part of~\cite{He2007}) is recorded in daylight as the methods usually become unusable in the night when only headlights of vehicles are visible. Existing datasets usually evaluate only speed measurement error (with different statistics -- mean, deviation etc.) and some exceptions (see Table~\ref{tab:SOTADatasets}) evaluate also other tasks.

The existing evaluation of algorithms should be considered insufficient as existing works use a small number of observed vehicles and scenes. Also, for GPS and speedometer, the ground truth is imprecise as in our evaluation GPS has mean error over 2\,\% and speedometer reports higher speed then the actual. Therefore, we created our novel dataset with precise ground truth and \CarsTotalCount of vehicles with ground truth speed. It is also possible to evaluate other camera calibration aspects such as calibration error and distance measurement on the road plane with the computed scale. These two metrics can provide interesting insights into properties of camera calibration algorithms as they are needed and harnessed in the intelligent transportation surveillance.

\begin{figure*}[t!]
	\centering
	\includegraphics[width=0.49\linewidth]{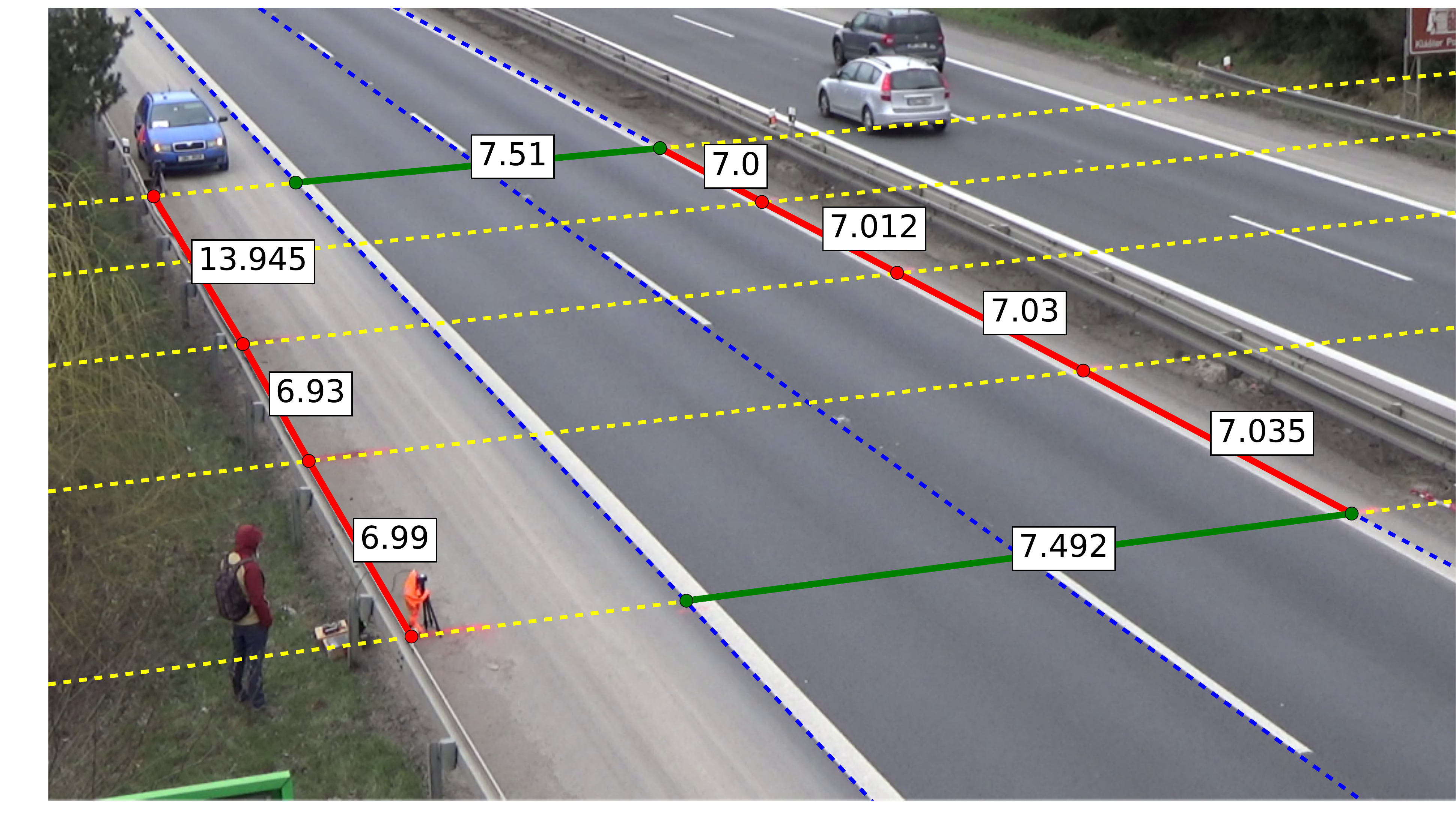}%
	\includegraphics[width=0.49\linewidth]{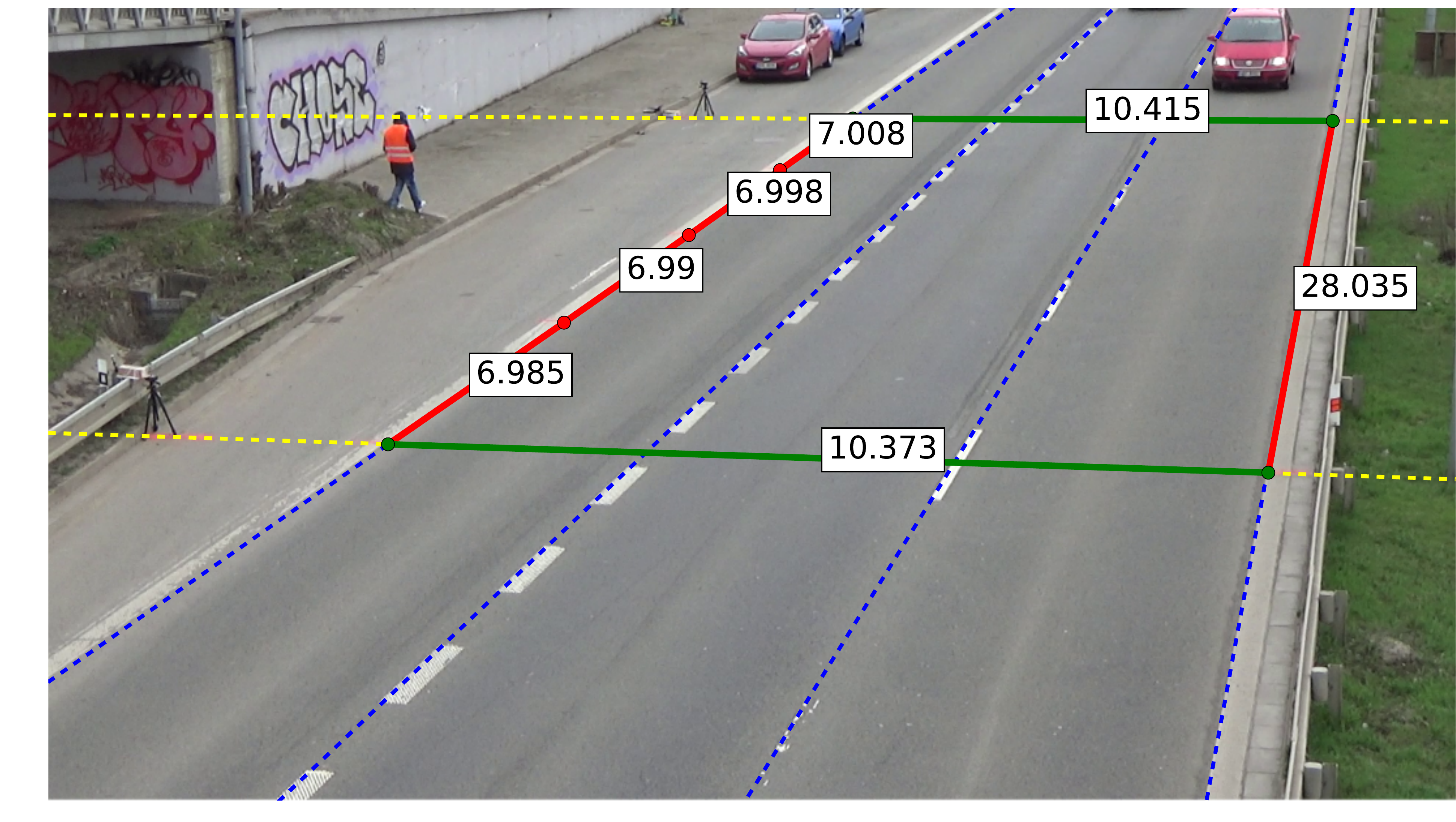}
	\caption{Markings and measured distances on the road plane. \textbf{Blue dashed lines} -- annotated lane dividing lines, \textbf{yellow dashed lines} -- measurement lines, \textbf{red line segments} -- measured distances towards the first vanishing point, \textbf{green line segments} -- measured distances towards the second vanishing point. Images with these annotations for all videos can be found in the supplementary material. Best viewed on screen.} \label{fig:Markings}
\end{figure*}

\section{Dataset Acquisition Methodology} \label{sec:Methodology}

We performed six recording sessions at different locations with free flow traffic.
For each session, we obtained three videos (approximately one hour long) from different positions by different video cameras (Panasonic HC-X920, Panasonic HDC-SD90, Sony Handycam HDR-PJ410). The videos were recorded in full-HD resolution and with 50 frames per second progressive scan. The recording setup is schematically shown in Figure \ref{fig:Setup} and an example of the scene is in Figure \ref{fig:Markings}.

Reference speed values of passing vehicles were obtained from a pair of experimental setups, containing a LIDAR (LaserAce\textsuperscript{\textregistered} IM HR 300), a GPS module (Leadtek LR9540D), and a PC.  These were placed on the side of the road perpendicular to the direction of traffic flow at a defined distance $D$ between them. It was important to place the lasers to the same height and parallel in the vertical and horizontal axes (see Figure \ref{fig:Setup}). This requirement guarantees that an incoming vehicle always disturbs the laser beams at the same point.

\begin{figure*}[t]
	\centering
	\def\svgwidth{0.8\linewidth}
	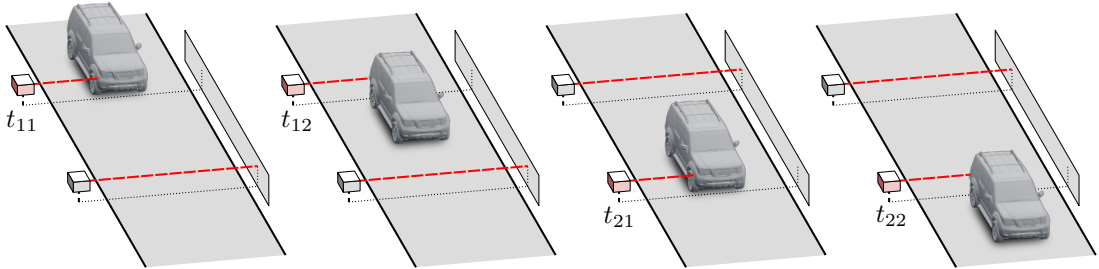
	\caption{
		The four phases of a passing car which are used for ground truth speed annotation. Best viewed on screen.
	}
	\label{fig:CarPassing}
\end{figure*}

The LIDAR works in the single shot mode (one laser pulse per range measurement). The sampling rate is 1\,kHz and maximal measurement range is 300\,m.
GPS receiver synchronizes times on PC using TIMEMARK signal (1 pulse per second with $1\,\mathrm{{\upmu}s}$ precision).
The data from each LIDAR and GPS module were recorded by the computer and each measurement was assigned with a high resolution timestamp obtained from the operating system.

\begin{figure*}[t]
	\centering
	\includegraphics[width=0.8\linewidth]{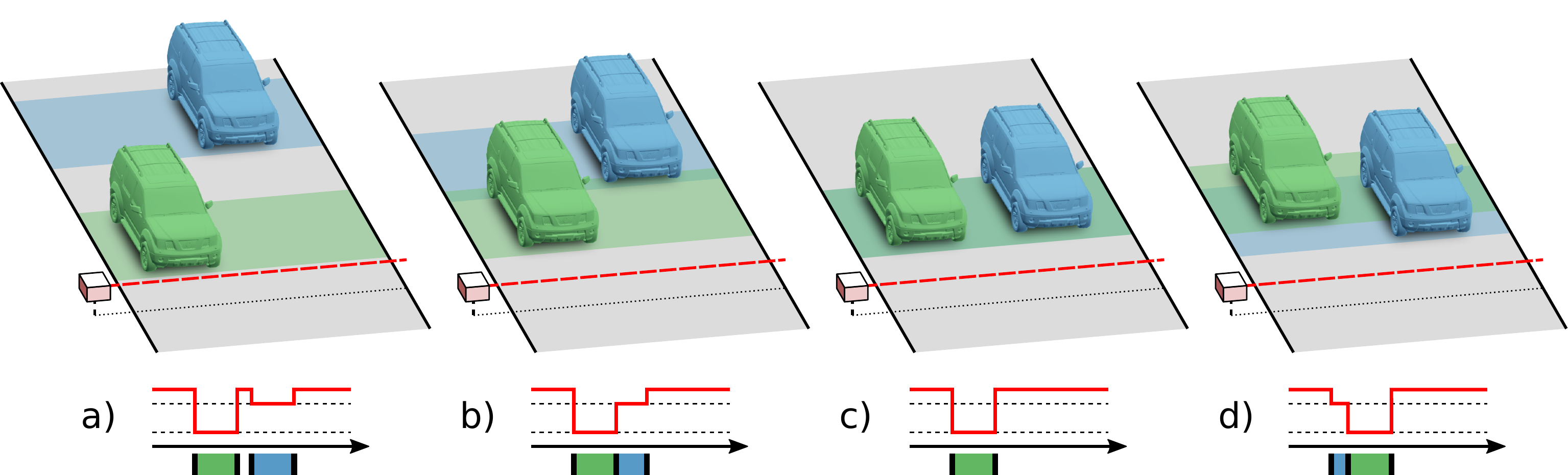}
	\caption{
		Possible variants of occlusion. \textbf{a)} the vehicles are not occluded at all, \textbf{b)} the closer vehicle is occluding the frontal part of the farther vehicle \textbf{c)} the farther vehicle is fully covered, \textbf{d)} the farther vehicle's rear part is covered. 
		The graphs below represent LIDAR responses with different levels for empty road, fast lane (top dashed line), and slower lane (bottom dashed lane). 
		See text for description of how all these situations are handled. Best viewed on screen.
	}
	\label{fig:Occlusions}
\end{figure*}

Distance logs from both LIDARs are pre-processed individually.
We search for timestamps $t_{xy}$ (see Figure~\ref{fig:CarPassing}) which correspond to car entering/leaving first/second laser beam. And each excitation is assigned with the lane based on the measured distance from the LIDAR.
Excitations generated by the same car on the first and second LIDAR need to be matched.
The matching is based on the correspondence of lane with limits on speed and acceleration of cars.
We calculate immediate speed when entering the first laser $v_{11}$ (at the time $t_{11}$), length of the vehicle $L$, and its average acceleration $a$ over measured span of known length $D$ by the following set of equations:
\begin{eqnarray}
	v_{11} + \frac{1}{2}a(t_{12}-t_{11}) = \frac{L}{t_{12}-t_{11}} \\
	v_{21} + \frac{1}{2}a(t_{22}-t_{21}) = \frac{L}{t_{22}-t_{21}} \\
	v_{11} + \frac{1}{2}a(t_{22}-t_{11}) = \frac{D+L}{t_{22}-t_{11}}
\end{eqnarray}
Then, it is possible to compute immediate speed at any point of the measured span.
Unfortunately, when a car is partially occluded by another vehicle, the equations above cannot be used for the calculation (as some timestamps are unknown).
If at least timestamps $t_{11}$ and $t_{21}$ are known, the average speed can be computed as
\begin{eqnarray}
	v_{avg} = \frac{D}{t_{21}-t_{11}}.
\end{eqnarray}

\begin{figure*}
	\centering
	\includegraphics[width=0.16\linewidth]{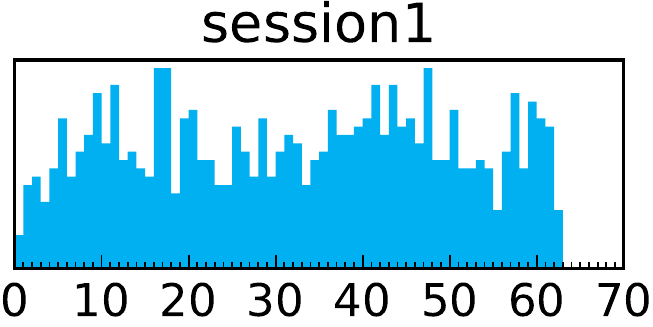}
	\includegraphics[width=0.16\linewidth]{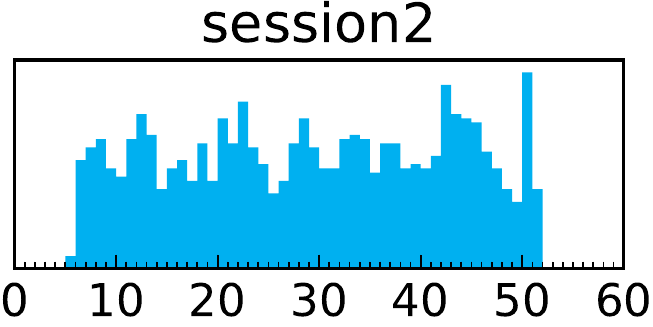}
	\includegraphics[width=0.16\linewidth]{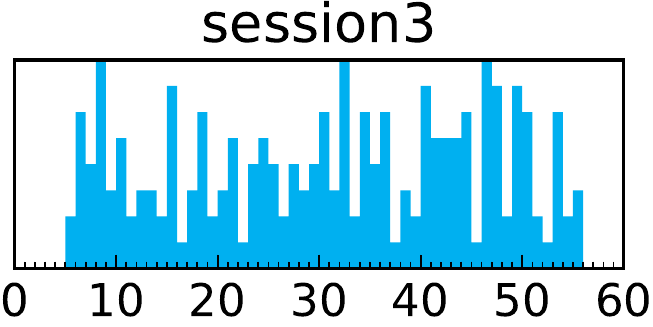}
	\includegraphics[width=0.16\linewidth]{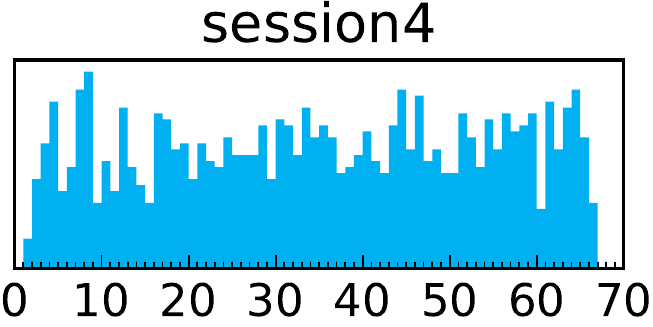}
	\includegraphics[width=0.16\linewidth]{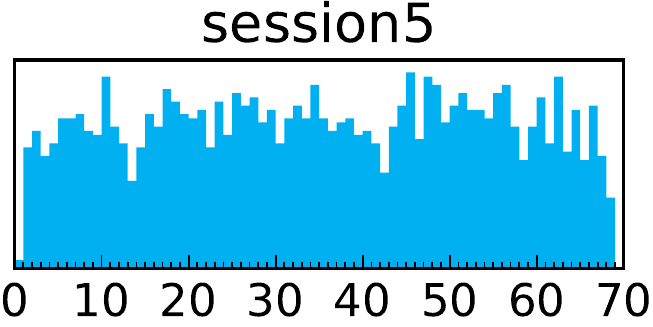}
	\includegraphics[width=0.16\linewidth]{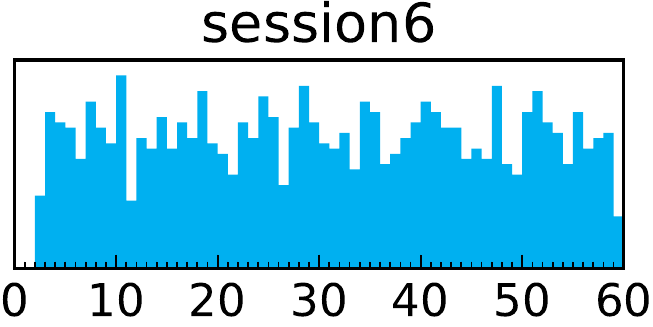}\\
	\includegraphics[width=0.16\linewidth]{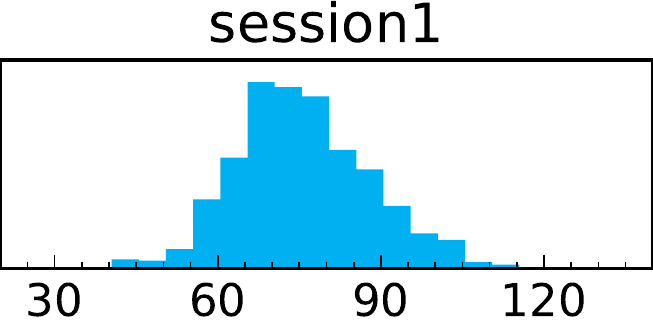}
	\includegraphics[width=0.16\linewidth]{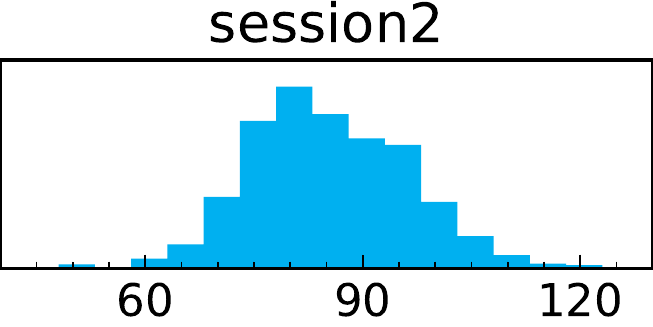}
	\includegraphics[width=0.16\linewidth]{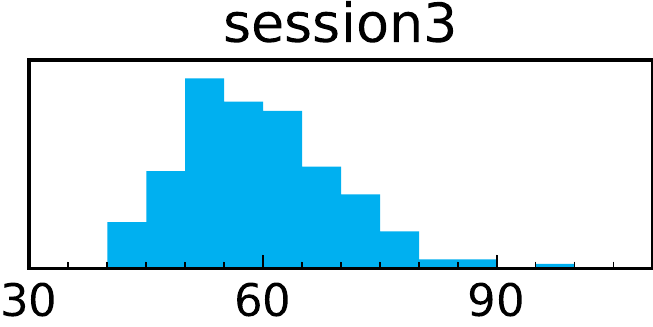}
	\includegraphics[width=0.16\linewidth]{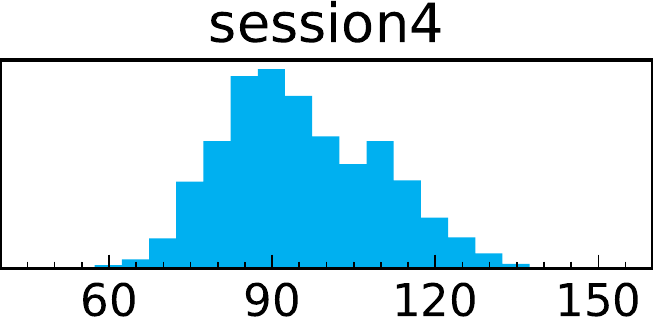}
	\includegraphics[width=0.16\linewidth]{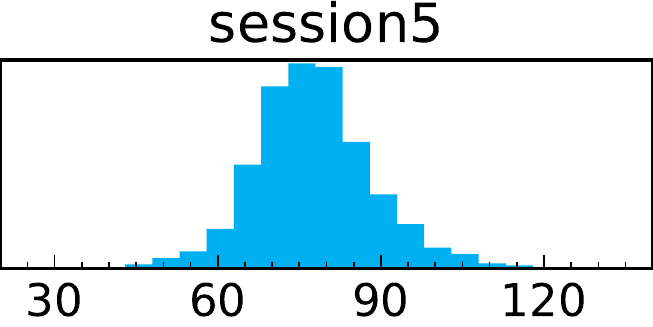}
	\includegraphics[width=0.16\linewidth]{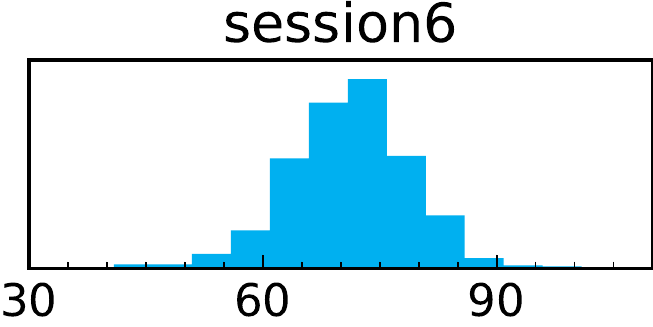}\\
	\includegraphics[width=0.16\linewidth]{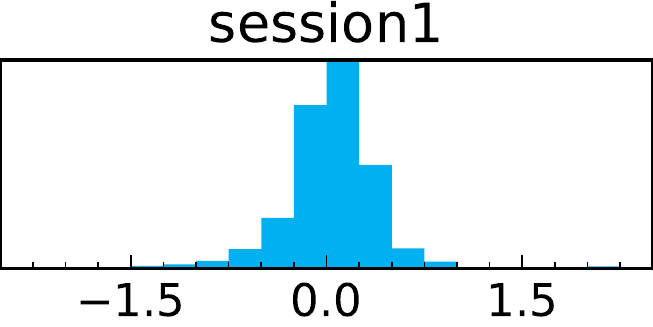}
	\includegraphics[width=0.16\linewidth]{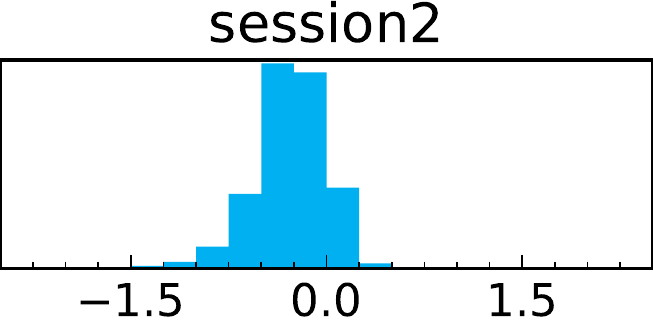}
	\includegraphics[width=0.16\linewidth]{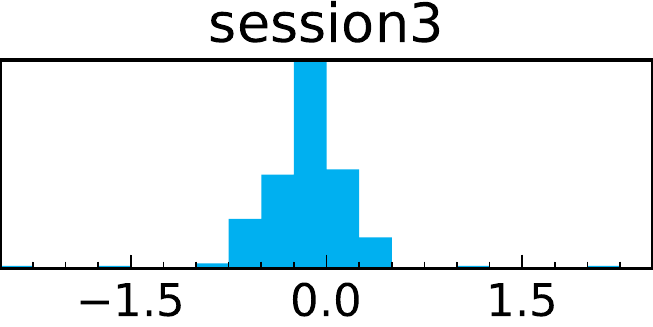}
	\includegraphics[width=0.16\linewidth]{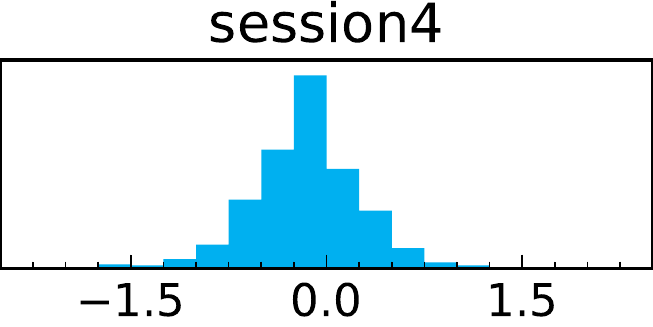}
	\includegraphics[width=0.16\linewidth]{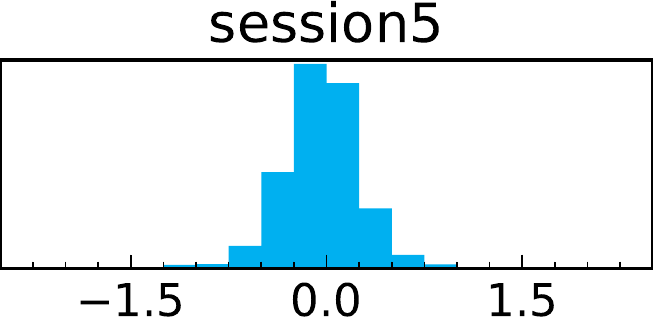}
	\includegraphics[width=0.16\linewidth]{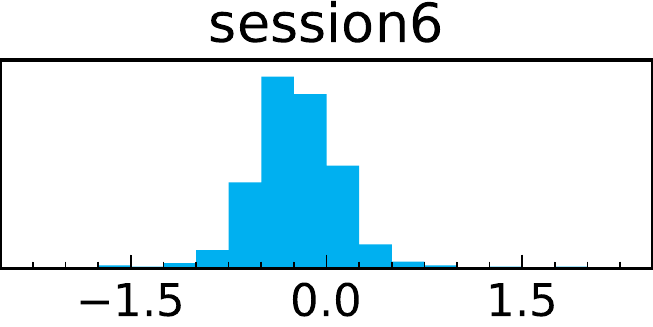}\\
	\caption{
		\textbf{top:} Histograms of density of vehicles for each session; one minute granularity in $x$-axis.
		\textbf{middle:} Ground truth speeds measured by the LIDAR setup (Section \ref{sec:Methodology}); speed in km/h on $x$-axis.
		\textbf{bottom:} Ground truth accelerations; $m/s^2$ on the $x$-axis.
	}
	\label{fig:Histograms}
\end{figure*}

As we are using LIDARs (instead of e.g. optical gates), we are able to detect situations when a vehicle is partially occluded by a closer vehicle using the measured distances by the LIDARs. See Figure \ref{fig:Occlusions} for examples of all possible occlusion types. There are several possibilities of occlusion on the pair of LIDARs:
\begin{enumerate}
	\item Occlusion situations on both the LIDARs are either \textbf{a)} or \textbf{d)} -- in these situations we are able to detect that there is a occluded vehicle and measure their speed. 
	\item Occlusion situations on at least one LIDAR is of type \textbf{b)} or \textbf{c)} -- we are able to detect that there is a second ``shadowed'' vehicle; the speed measurement is not reliable and the second vehicle is omitted from the dataset and evaluation. 
	\item Occlusion situations on \textbf{both} LIDARs are \textbf{c)} -- the second ``shadowed'' vehicle cannot even be detected. This situation is very unlikely, as the vehicle in the fast lane would have to be smaller, precisely aligned, and maintain the same speed as the closer vehicle. 
\end{enumerate}
In summary, we either measure the speed accurately, or we know that the speed measurement is not precise and we ignore such a measurement. Therefore, besides the 20\,865 vehicles with precise ground truth speed, the dataset contains 2\,779 instances of vehicles which are marked as invalid for speed measurement evaluation.

We also performed manual verification of the matched timestamps $t_{11}$ and $t_{21}$ by checking that they correspond to the same vehicle in the video.

\subsection{Accuracy of the Acquired Dataset}
The distance $D$ between LIDARs is 28 meters (21 meters in one case), and the LIDARs have 1\,kHz sampling rate.
The actual value of $D$ for every recording session was measured by handheld laser distance meter, and we assume that upper bound of the distance measurement error is $e_d = 0.05\,\mathrm{m}$. Time measurement error caused by improper synchronization of LIDARs is at most $e_t = 1\,\mathrm{ms}$. Both, $e_d$ and $e_t$ are exaggerated and in reality they are lower.
The upper bound of speed measurement error $E_r$ (relative) and $E_a$ (absolute) for the given speed $v$ can be computed as:

\noindent\begin{minipage}{.35\linewidth}
\begin{align*}
E_r &=  \frac{e_d + e_t \cdot v}{D}\\
E_a &=  E_r \cdot v
\end{align*}
\end{minipage}
\qquad
\begin{minipage}{.60\linewidth}
	\centering
	\includegraphics[width=\textwidth]{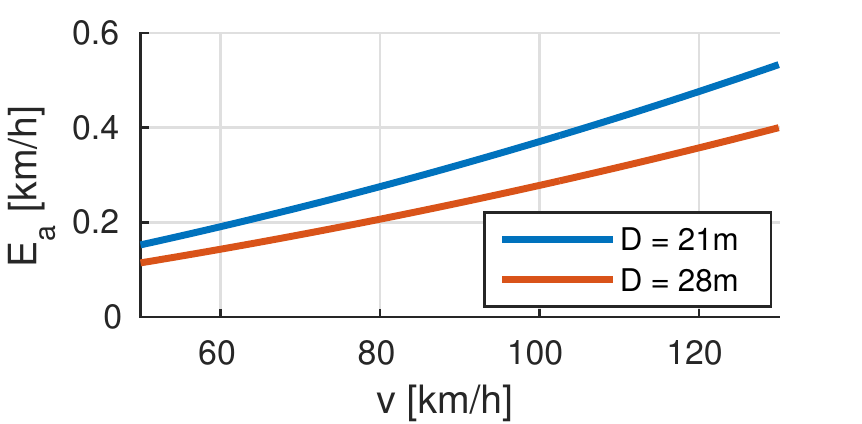}
\end{minipage}\\[2pt]

For a vehicle going $v = 20\,\mathrm{m/s}$ ($72\,\mathrm{km/h}$), the resulting maximum possible errors are $E_r = 0.25\,\%$ and $E_a = 0.05\,\mathrm{m/s}$ ($0.18\,\mathrm{km/h}$). We consider these errors to be small enough as the errors of the methods presented in Section \ref{sec:Experiments} are much higher than this error of measurement.


\begin{table}
	\centering
	\caption{Numbers of vehicle passes with known ground truth for each video.} \label{tab:ObservedVehicles}
	\setlength{\tabwidth}{0.75\linewidth}
	\small
	\begin{tabular}{c r r r }
		\toprule
		& \textbf{left} & \textbf{center} & \textbf{right}\\
		\midrule
		\textbf{session1} & 854 & 848 & 849\\
		\textbf{session2} & 1\,163 & 1\,258 & 1\,583\\
		\textbf{session3} & 193 & 193 & 193\\
		\textbf{session4} & 1\,188 & 1\,192 & 1\,177\\
		\textbf{session5} & 2\,021 & 2\,027 & 2\,030\\
		\textbf{session6} & 1\,358 & 1\,353 & 1\,358\\
		\midrule
		\textbf{TOTAL} & \multicolumn{3}{c}{\CarsTotalCount}\\
		\bottomrule
	\end{tabular}
	
\end{table}

\section{Dataset Statistics and Evaluation Protocol}
The dataset consists of 18 videos (6 sessions on different locations, 3 videos from different angles for each location) and there is totally \CarsTotalCount vehicles with known ground truth speed.

To provide statistics about the dataset we report the total number of cars with ground truth speed for each video in Table \ref{tab:ObservedVehicles}. We also report histograms of speeds, accelerations, and traffic density in Figure~\ref{fig:Histograms}.

The dataset is (to our knowledge) by far larger than other datasets serving similar purpose reported in the literature. It covers views typical for traffic surveillance from arbitrary cameras. It provides high quality videos with various traffic conditions (low traffic in Session 3, high traffic in Sessions 5 and 6). However, it is quite limited in lighting and weather conditions. Almost all videos were taken in cloudy weather (except some parts of Session 3) with no distracting phenomena (fog, rain, etc.).

\subsection{Evaluation Protocol}

For future comparison of methods, we provide an evaluation script%
\footnote{The evaluation code is available together with the dataset at \url{https://medusa.fit.vutbr.cz/traffic}}
which automatically evaluates all the used metrics.
It requires two vanishing points of the road plane, principal point of the camera and scale of the scene as the calibration parameters. Then, the systems are supposed to report for each observed vehicle a track of one arbitrary reference point on the road plane (frame numbers + image coordinates). In our case, the point is obtained by the constructed 3D bounding boxes (see Figure~\ref{fig:MethodScheme}). The point must be on the road plane for proper projection; however it can be any point on the road plane which the authors are able to localize -- it is not necessary to use the 3D bounding boxes.

To compare vehicles with the ground truth, we match the time
when a vehicle passed the measurement line to the time reported by LIDAR and the lane in which the vehicle is. As the vehicles are sometimes not tracked correctly and the tracking can be lost, we extrapolate the vehicle trajectory in order to get the correct time.

For each vehicle, we calculate tentative speeds between the positions $K$ frames apart (approximately 0.1\,s, $K=5$ for 50 fps video) by projecting the image point coordinates to the road plane using the provided calibration. The resulting speed is then median of the tentative speeds. We found out that this method is more robust than measuring the full section speed due to possible tracking errors.

The computation of distance between two points $p_1$ and $p_2$ is schematically shown in Figure \ref{fig:Projection}  with general model for traffic surveillance camera and it is described in detail in the supplementary material.

As methods may require different training sets we define three train/test splits. Split~\textbf{A} uses all videos for testing, split~\textbf{B} has Session 1 and Session 2 reserved for training, and finally, split~\textbf{C} has Session 1, Session 2, and Session 3 for training. Whenever it is possible, the results should be reported on the splitting with the lowest number of training sessions.

\section{Experiments} \label{sec:Experiments}

On the above described dataset we evaluate recent method~\cite{Dubska2014} described in Section~\ref{sec:method}; we use this method for the evaluation because it works fully automatically (contrary to \cite{Maduro2008,Nurhadiyatna2013,Sina2013,Luvizon2014}) and it is not limited to some viewpoints (contrary to \cite{Dailey2000,Grammatikopoulos2005,Nurhadiyatna2013,Lan2014,Do2015}).  
The method is able to automatically recover camera calibration and scene scale. 
However, our dataset provides  data in the form of measured distances on the road plane usable for computing camera calibration (vanishing points and scene scale). Therefore, we also report the performance of the semi-automatic variants of the method.

We defined labels for different camera calibrations (the vehicle detection and tracking stays the same for all of the methods):\\
\textbf{FullACC} \cite{Dubska2014} -- unmodified system from \cite{Dubska2014}, as it is Fully Automatic Camera Calibration.\\
\textbf{OptScale},\textbf{OptScaleVP2} -- Keep calibration (vanishing points) from FullACC and calculate optimal scale using lengths in direction to VP1 (OptScale) or VP2 (OptScaleVP2). The scale is computed as a mean of scale values obtained from the distance measurements on the road.\\
\textbf{OptCalib},\textbf{OptCalibVP2} -- The first vanishing point is kept from the FullACC. And as the second vanishing point is selected a point which minimizes the calibration error (see Section~\ref{sec:ExperimentsCalibError}). The minimization is done by a grid search in space of feasible vanishing points. The scale is computed the same way as for OptScale and OptScaleVP2.  

On these five variants we report the calibration error, distance measurement error, and speed measurement error.
The speed error is additionally compared to the GPS speed. The evaluation in this chapter is done on split \textbf{A}, as the method does not require any training so we can use all the videos for evaluation. Evaluation for each video separately can be done directly from the published dataset as we included also the results.

As the camera calibration algorithm is the most sensitive part of a speed measurement system, we provide also an evaluation of the calibration itself based on two detected vanishing points and evaluation of distance measurement accuracy on the road plane. These two metrics are also important as they compare directly the camera calibrations without any influence of vehicle detection and tracking.

For each presented evaluation metric we propose to report mean, median and 95 percentile and where it is possible (speed measurement error), we also report absolute and relative cumulative histograms of errors. These statics correspond to used evaluation metrics and methods shown in Table~\ref{tab:SOTADatasets}.

\subsection{Calibration Error} \label{sec:ExperimentsCalibError}
The first evaluation experiment is focused on the calibration itself (detected two vanishing points) excluding the scale. We measure the ratio between every pair of distances measured on the road plane (see Figure~\ref{fig:Markings}) and compare it with the ratio of the dimensions measured using the calibration. The scale is therefore omitted from this evaluation and the results depend only on the two vanishing points.

For each system, we measure mean, median and 95 percentile error for both absolute units ($err = |r_{gt} - r_{m}|$) and relative units $(err = |r_{gt} - r_{m}|/r_{gt}\cdot 100\%$), where $r_{gt}$ denotes the ground truth ratio, and $r_{m}$ represents the measured ratio. This computation of absolute and relative error is used also in Sections \ref{sec:DistanceMeasurementEvaluation} and \ref{sec:SpeedMeasurementEvaluation}.
The results can be found in Table \ref{tab:CalibrationError}. As there are two groups of methods (\textbf{FullACC}+\textbf{OptScale}+\textbf{OptScaleVP2} and \textbf{OptCalib}+\textbf{OptCalibVP2}) which share the calibration (vanishing points) and are differentiated only in the scale which is not used in this experiment, the results are the same for methods within each group.

The results show that the camera calibration automatically obtained by \cite{Dubska2014} is far from perfect. The biggest error is caused by inaccurate localization of the second vanishing point (VP2). Thus the lengths in the direction to VP2 (i.e. widths of vehicles) are unreliable for scale computation.

\begin{table}
	\centering
	\caption{Errors for distance ratios (see text for details). The first row for each calibration method contains \abso{absolute errors} and the \proc{relative errors in percents} are in the second row.}  \label{tab:CalibrationError}
	\setlength{\tabwidth}{0.93\linewidth}
	\begin{tabular}{l r r r }
		\toprule
		\textbf{system} & \textbf{mean} & \textbf{median} & \textbf{95\,\%} \\
		\midrule

		\multirow{2}{*}{FullACC \cite{Dubska2014}, OptScale, OptScaleVP2} & \abso{0.15} & \abso{0.04} & \abso{0.56}\\
		& \proc{10.89} & \proc{4.52} & \proc{40.24}\\[-\jot]			    \multicolumn{4}{@{}c@{}}{\makebox[\tabwidth]{\dashrule[lightgray]}} \\[-\jot]

		\multirow{2}{*}{OptCalib, OptCalibVP2} & \abso{0.03} & \abso{0.01} & \abso{0.09}\\
		& \proc{2.62} & \proc{1.58} & \proc{8.79}\\

		\bottomrule
	\end{tabular}
\end{table}

\subsection{Distance Measurement Error} \label{sec:DistanceMeasurementEvaluation}


\begin{table}
	\centering
	\caption{Errors for distance measurement \textbf{towards the first vanishing point} (see text for details). The first row for each calibration method contains \abso{absolute errors in meters} and the \proc{relative errors in percents} are in the second row.} \label{tab:DistanceMeasurementsVP1}
	\setlength{\tabwidth}{0.60\linewidth}
	\begin{tabular}{l r r r }
		\toprule
		\textbf{system} & \textbf{mean} & \textbf{median} & \textbf{95\,\%} \\
		\midrule

		\multirow{2}{*}{FullACC \cite{Dubska2014}} & \abso{1.41} & \abso{1.06} & \abso{4.45}\\
		& \proc{12.32} & \proc{12.00} & \proc{25.13}\\[-\jot]			    \multicolumn{4}{@{}c@{}}{\makebox[\tabwidth]{\dashrule[lightgray]}} \\[-\jot]

		\multirow{2}{*}{OptScale} & \abso{0.23} & \abso{0.13} & \abso{1.18}\\
		& \proc{1.94} & \proc{1.45} & \proc{5.05}\\[-\jot]			    \multicolumn{4}{@{}c@{}}{\makebox[\tabwidth]{\dashrule[lightgray]}} \\[-\jot]

		\multirow{2}{*}{OptScaleVP2} & \abso{2.61} & \abso{1.60} & \abso{8.53}\\
		& \proc{21.86} & \proc{20.21} & \proc{57.62}\\[-\jot]			    \multicolumn{4}{@{}c@{}}{\makebox[\tabwidth]{\dashrule[lightgray]}} \\[-\jot]

		\multirow{2}{*}{OptCalib} & \abso{0.14} & \abso{0.09} & \abso{0.41}\\
		& \proc{1.43} & \proc{0.92} & \proc{3.56}\\[-\jot]			    \multicolumn{4}{@{}c@{}}{\makebox[\tabwidth]{\dashrule[lightgray]}} \\[-\jot]

		\multirow{2}{*}{OptCalibVP2} & \abso{0.34} & \abso{0.14} & \abso{1.74}\\
		& \proc{2.46} & \proc{1.54} & \proc{8.05}\\

		\bottomrule
	\end{tabular}
\end{table}

To evaluate the distance measurement including the scale, we carried out the next experiment where we compared ground truth distances on the road plane (see Figure \ref{fig:Markings}) and distances obtained using the camera calibration converted to meters using the scale.

We divided the experiment into two parts. The first one is focused only on distances towards the first vanishing point, as these are the most important for the speed measurement. The results can be found in Table~\ref{tab:DistanceMeasurementsVP1}.
In the second part of the experiment, we evaluated all the distances measured on the road plane and the results can be found in Table~\ref{tab:DistanceMeasurementsAll}.

\begin{table}[t]
	\centering
	\caption{Errors for \textbf{all} distance measurements (see text for details). The first row for each calibration method contains \abso{absolute errors in meters} and the \proc{relative errors in percents} are in the second row.} \label{tab:DistanceMeasurementsAll}
	\setlength{\tabwidth}{0.60\linewidth}
	\begin{tabular}{l r r r }
		\toprule
		\textbf{system} & \textbf{mean} & \textbf{median} & \textbf{95\,\%} \\
		\midrule

		\multirow{2}{*}{FullACC \cite{Dubska2014}} & \abso{1.30} & \abso{0.85} & \abso{4.47}\\
		& \proc{12.11} & \proc{10.91} & \proc{25.64}\\[-\jot]			    \multicolumn{4}{@{}c@{}}{\makebox[\tabwidth]{\dashrule[lightgray]}} \\[-\jot]

		\multirow{2}{*}{OptScale} & \abso{0.62} & \abso{0.17} & \abso{2.47}\\
		& \proc{6.83} & \proc{2.05} & \proc{32.04}\\[-\jot]			    \multicolumn{4}{@{}c@{}}{\makebox[\tabwidth]{\dashrule[lightgray]}} \\[-\jot]

		\multirow{2}{*}{OptScaleVP2} & \abso{1.98} & \abso{1.21} & \abso{6.41}\\
		& \proc{16.84} & \proc{12.94} & \proc{49.98}\\[-\jot]			    \multicolumn{4}{@{}c@{}}{\makebox[\tabwidth]{\dashrule[lightgray]}} \\[-\jot]

		\multirow{2}{*}{OptCalib} & \abso{0.14} & \abso{0.07} & \abso{0.60}\\
		& \proc{1.58} & \proc{0.68} & \proc{5.24}\\[-\jot]			    \multicolumn{4}{@{}c@{}}{\makebox[\tabwidth]{\dashrule[lightgray]}} \\[-\jot]

		\multirow{2}{*}{OptCalibVP2} & \abso{0.30} & \abso{0.12} & \abso{1.12}\\
		& \proc{2.37} & \proc{1.44} & \proc{8.47}\\

		\bottomrule
	\end{tabular}

\end{table}

\begin{table}
	\centering
	\caption{Errors for speed measurements (see text for details). The first row for each calibration method contains \abso{absolute errors in km/h} and the \proc{relative errors in percents} are in the second row.} \label{tab:SpeedMeasurement}
	\setlength{\tabwidth}{0.60\linewidth}
	\begin{tabular}{l r r r }
		\toprule
		\textbf{system} & \textbf{mean} & \textbf{median} & \textbf{95\,\%} \\
		\midrule
		\multirow{2}{*}{GPS} & \abso{1.64} & \abso{1.19} & \abso{---}\\
		& \proc{2.18} & \proc{1.42} & \proc{---}\\[-\jot]			    \multicolumn{4}{@{}c@{}}{\makebox[\tabwidth]{\dashrule[lightgray]}} \\[-\jot]
		
		\multirow{2}{*}{RADAR} & \abso{1.07} & \abso{0.89} & \abso{2.69}\\
		& \proc{1.33} & \proc{1.14} & \proc{3.23}\\\midrule

		\multirow{2}{*}{FullACC \cite{Dubska2014}} & \abso{8.59} & \abso{8.45} & \abso{17.14}\\
		& \proc{10.89} & \proc{11.41} & \proc{19.84}\\[-\jot]			    \multicolumn{4}{@{}c@{}}{\makebox[\tabwidth]{\dashrule[lightgray]}} \\[-\jot]
		
		\multirow{2}{*}{OptScale} & \abso{1.71} & \abso{1.17} & \abso{4.69}\\
		& \proc{2.13} & \proc{1.51} & \proc{5.56}\\[-\jot]			    \multicolumn{4}{@{}c@{}}{\makebox[\tabwidth]{\dashrule[lightgray]}} \\[-\jot]
		
		\multirow{2}{*}{OptScaleVP2} & \abso{15.66} & \abso{13.09} & \abso{47.86}\\
		& \proc{19.83} & \proc{17.51} & \proc{59.25}\\[-\jot]			    \multicolumn{4}{@{}c@{}}{\makebox[\tabwidth]{\dashrule[lightgray]}} \\[-\jot]
		
		\multirow{2}{*}{OptCalib} & \abso{1.43} & \abso{0.83} & \abso{3.89}\\
		& \proc{1.81} & \proc{1.05} & \proc{5.07}\\[-\jot]			    \multicolumn{4}{@{}c@{}}{\makebox[\tabwidth]{\dashrule[lightgray]}} \\[-\jot]
		
		\multirow{2}{*}{OptCalibVP2} & \abso{2.43} & \abso{1.40} & \abso{6.66}\\
		& \proc{3.08} & \proc{1.76} & \proc{8.00}\\
		
		\bottomrule
	\end{tabular}
\end{table}

The results in Table \ref{tab:DistanceMeasurementsVP1} show many different interesting aspects of the algorithm. The first one is that if we use the original calibration and use scale computed from distances towards the first vanishing point (\textbf{OptScale}) it improves the results significantly. However, when we use the same calibration and scale computed from distances towards the second vanishing point (\textbf{OptScaleVP2}), then the results are even worse than the original ones. This implies that the error is in the localization of the second vanishing point. The table also shows that in a situation when we use the correctly localized second vanishing point, the results significantly improve for both \textbf{OptCalib} and \textbf{OptCalibVP2} relative to \textbf{OptScale} and \textbf{OptScaleVP2}.

\begin{figure*}[t]
	\centering
	\includegraphics[width=0.48\linewidth]{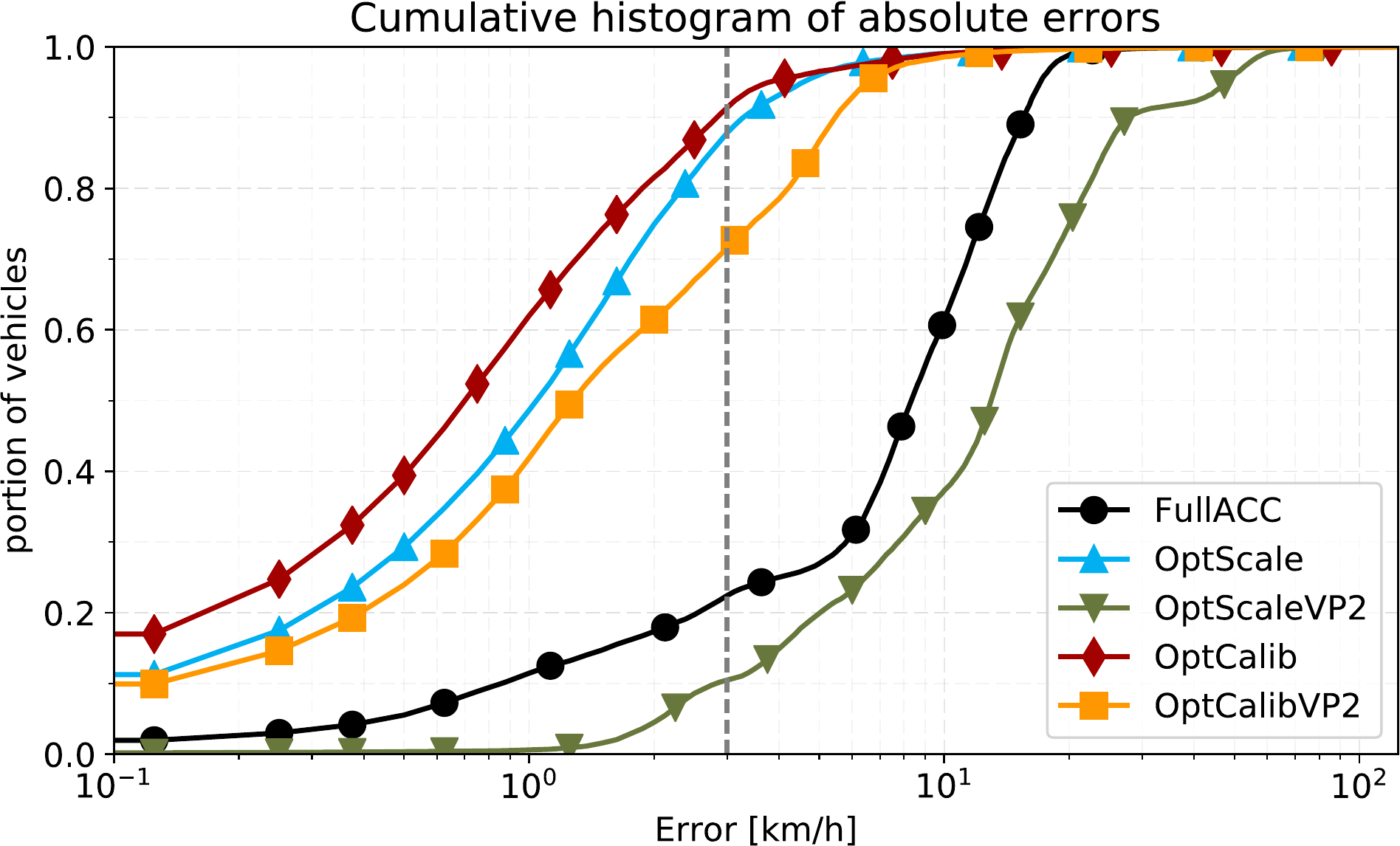}\qquad
	\includegraphics[width=0.48\linewidth]{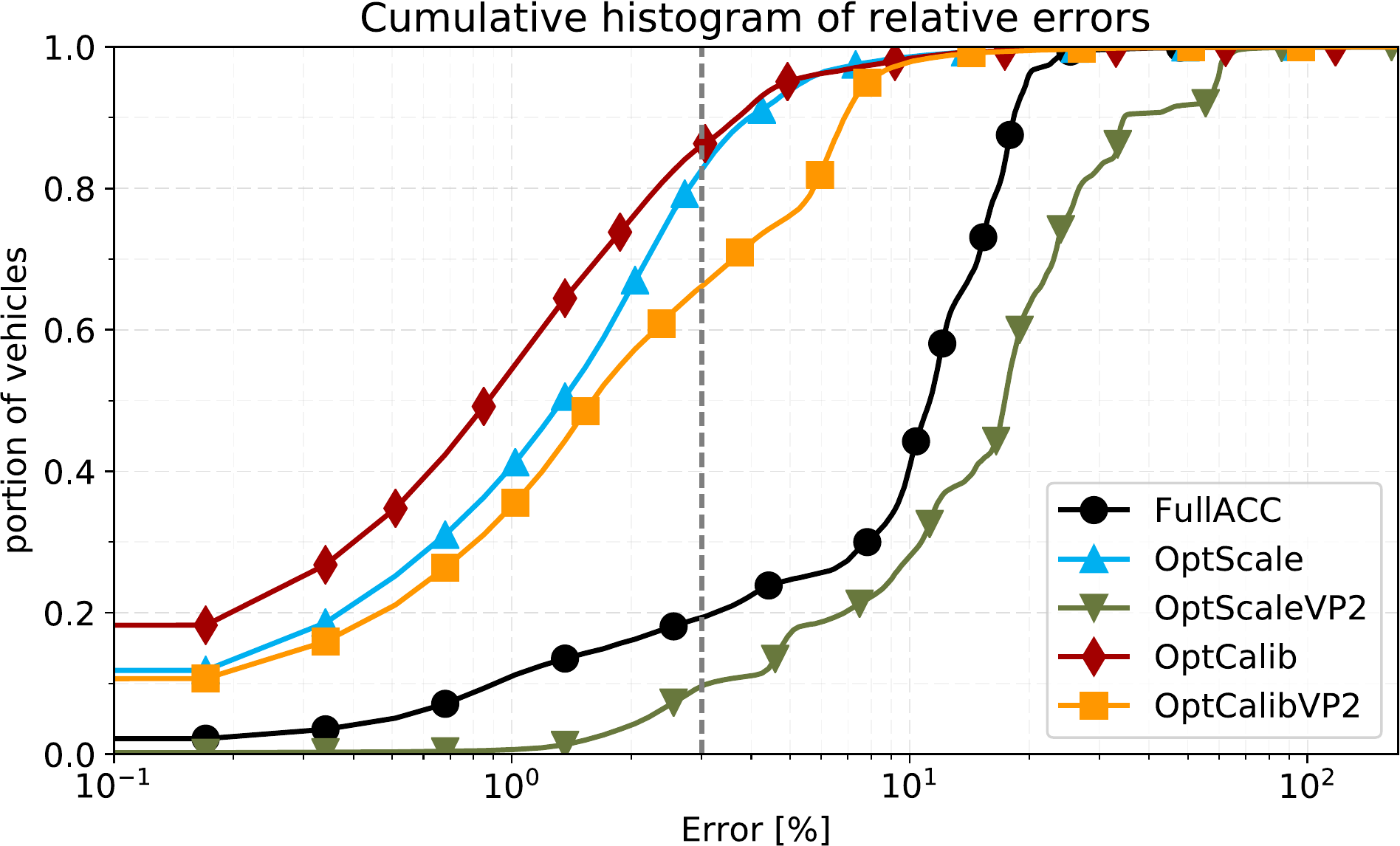}
	\caption{Cumulative histograms of distribution of errors. The dashed vertical line represents 3\,km/h (or 3\,\%) threshold. See text for details. (Line markers represent every 1500$^\mathrm{th}$ data sample.)} \label{fig:ErrorsCumHistograms}
\end{figure*}

Results in Table \ref{tab:DistanceMeasurementsAll} support the hypothesis that the second vanishing point is incorrectly detected by the method \cite{Dubska2014} in some cases, as the distance measurement performance significantly drops for \textbf{OptScale} when distances towards the second vanishing point are added to the evaluation (Table~\ref{tab:DistanceMeasurementsVP1} vs Table \ref{tab:DistanceMeasurementsAll}). Also, it shows that \textbf{OptCalib} and \textbf{OptCalibVP2} are not affected by this.
Also, as we expected, the performance of \textbf{OptScaleVP2} increases when the dimensions towards the second vanishing point are added to the evaluation.
The original system \textbf{FullACC} does not have this significant drop in performance as the scale is determined either from widths, lengths, or heights; so it can be matched to the the correct scale for measurements of the distances towards the second vanishing point.

\subsection{Speed Measurement Error} \label{sec:SpeedMeasurementEvaluation}
For the speed measurement task itself, we evaluate mainly the error between the ground truth speed and the measured one. This metric does not include statistics about incorrectly detected and tracked vehicles.

The results on the evaluation videos can be found in Table~\ref{tab:SpeedMeasurement} and cumulative histograms of errors are shown in Figure~\ref{fig:ErrorsCumHistograms}. To compare the results with another non-visual speed measurement method, we also drove a car with GPS system with enabled raw logging to be seen in the recordings multiple times and computed the speed of these observations from the GPS logs by averaging over a longer period.  We used nVidia Shield tablets as our GPS loggers in ``Device only'' mode (GPS localization ON, Wi-Fi and GSM localization OFF) and logged the GPS data in NMEA format using a standard logging application available in the application store.
We process offline RMC messages, distance and velocity between each two following points are computed using Haversine formula described by Robusto~\cite{Robusto1957}. For each evaluation video we have approximately 20 passes with the GPS speed measured.

The results in Table \ref{tab:SpeedMeasurement} and Figure \ref{fig:ErrorsCumHistograms} show that the systems \textbf{OptCalib}, \textbf{OptScale} and \textbf{OptCalibVP2} work relatively well (with \textbf{OptCalib} being the best). Also, the results show that the \textbf{FullACC} has lower errors than \textbf{OptScaleVP2} implying that the biggest problem is in bad localization of the second vanishing point as was described in Section~\ref{sec:DistanceMeasurementEvaluation}. The results also show that when the original vanishing points from \textbf{FullACC} are used and the scale is computed to optimize the error in distance measurement towards the first vanishing point (\textbf{OptScale}), the performance increase significantly. Also, when the second vanishing point is correctly localized (\textbf{OptCalib} and \textbf{OptCalibVP2}) the results improves furthermore.

Table \ref{tab:SpeedMeasurement} also shows that the optimal system \textbf{OptCalib} outperforms the results obtained from the GPS speed measurements. The 95 percentile is not reported for the GPS as there is a much smaller number of measurements for the GPS than in the other cases; thus the numbers are not comparable. Also, we evaluated the speed measurement done by RADARs (2D~microwave FM-CW radar module RFbeam K-MC4 operating in K-band) and we used one RADAR for each lane for each evaluation session in order to compare them with results obtained from the methods using video; see Table \ref{tab:SpeedMeasurement}.

We also evaluated the number of false positives per minute of video (9.745) and recall (0.872) for all the videos. In the case of vehicle counting, false positives represent reported vehicle tracks which are not present in the dataset; recall denotes the fraction of correctly matched ground truth vehicle tracks with reported vehicle tracks. The results are the same for all the systems as they share the vehicle detection and tracking part and they are only different in the camera calibration. The false positives are caused mainly by lost tracking and re-initialization. Another important drawback of the current method is that the motion is not detected correctly in some cases and the motion mask is divided into several contours.

Although the systems with some manual calibration (\textbf{OptScale} and \textbf{OptCalib}) have relatively low speed measurement errors in comparison with GPS and RADAR, the fully automatic system still has too large errors and the automatic traffic surveillance camera calibration methods need improvement.

\section{Conclusions}

We collected and processed a dataset for evaluation of purely visual speed measurement by a single monocular camera.
Cameras are becoming ubiquitous and a considerable portion of them observe traffic.  By providing this dataset we intend to encourage research of fully automatic traffic camera calibration methods, which could be used for mining valuable automatic traffic surveillance data from existing and new camera infrastructure.

On the collected data, we evaluated an approach which is both fully automatic and can process virtually arbitrary views.  The evaluation shows its weaknesses (localization of the VP2 and scale inference), which can encourage further research in this area, which we will focus on.  The measurements also established a first baseline to be outperformed by future works.

\section*{Acknowledgment}
This work was supported by The Ministry of Education, Youth and Sports of the Czech Republic from the National Programme of Sustainability (NPU II); project IT4Innovations excellence in science -- LQ1602.

{\small
	\bibliographystyle{IEEEtran}
	\bibliography{2016-ITS-BrnoCompSpeed-bibliography}
}
\vspace{-1cm}
\begin{IEEEbiography}[{\includegraphics[width=1in,height=1.25in,keepaspectratio]{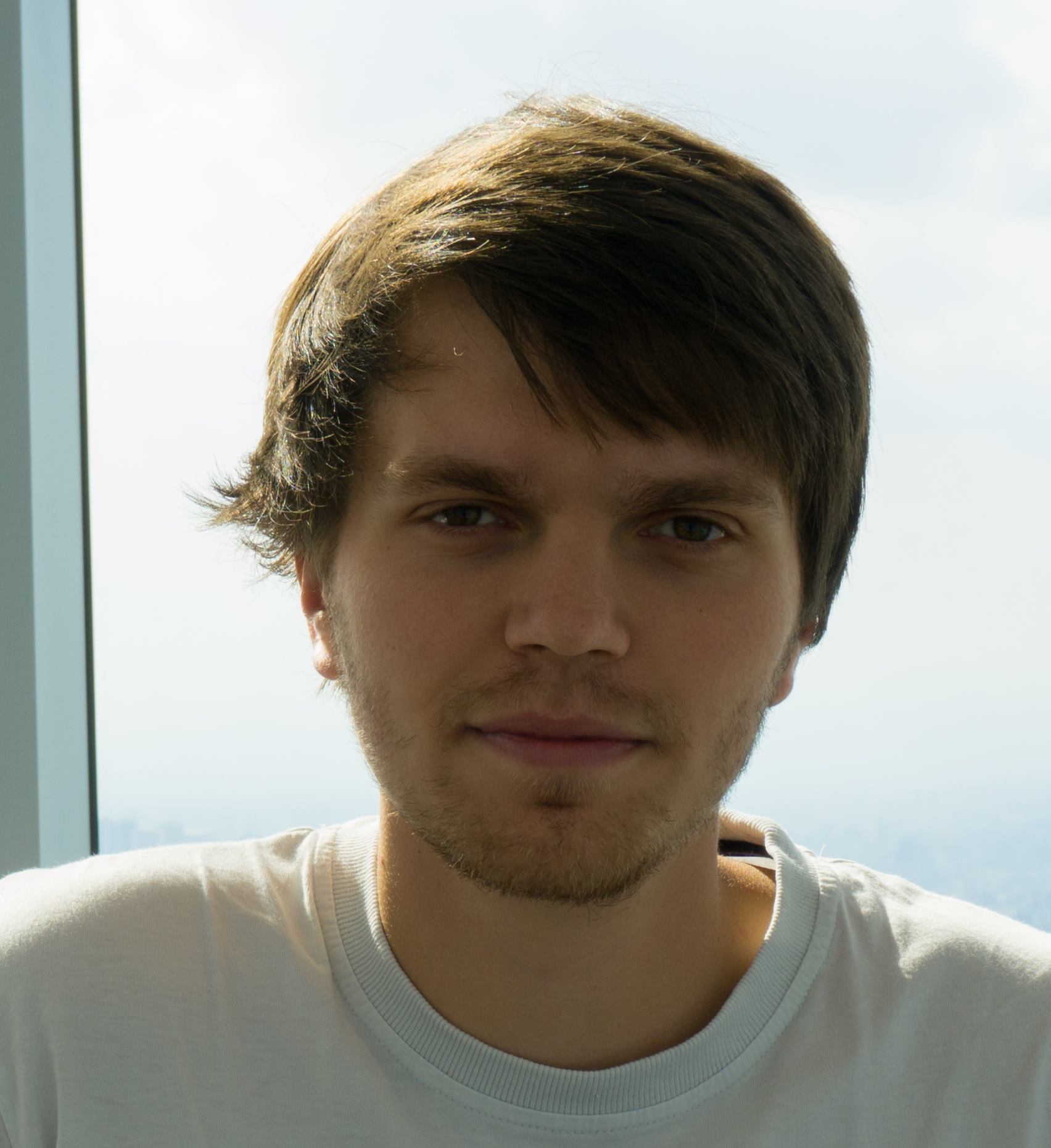}}]{Jakub Sochor}
received the M.S. degree from
Brno University of Technology (BUT), Brno, Czech
Republic. He is currently working toward the Ph.D.
degree in the Department of Computer Graphics
and Multimedia, Faculty of Information Technology,
BUT.
His research focuses on computer vision, particularly
traffic surveillance -- fine-grained recognition of vehicles and automatic speed measurement.
\end{IEEEbiography}
\vspace{-0.35cm}
\begin{IEEEbiography}[{\includegraphics[width=1in,height=1.25in,clip,keepaspectratio]{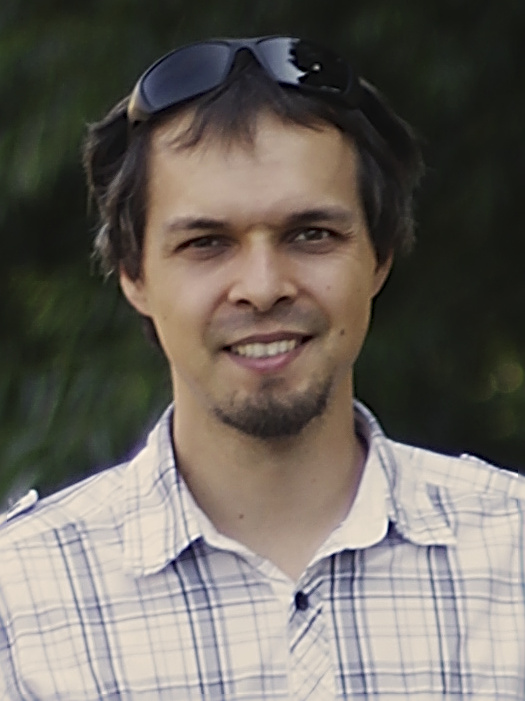}}]{Roman Juránek}
received his PhD from Faculty of Information Technology, Brno University of Technology, Czech Republic, where he works as research scientist. Currently he focuses on computer vision in traffic surveillance. His research interests include machine learning, image processing, computer vision and fast object detection.
\end{IEEEbiography}
\vspace{-0.35cm}
\begin{IEEEbiography}[{\includegraphics[width=1in,height=1.25in,clip,keepaspectratio]{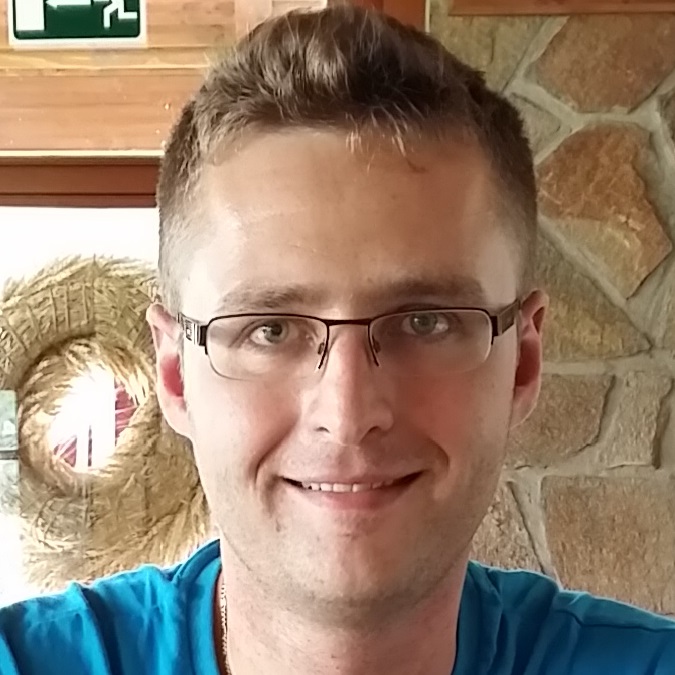}}]{Jakub Špaňhel}
received his M.S. degree from Faculty of Information Technology,
Brno University of Technology (BUT), Czech
Republic. He is currently working toward the Ph.D.
degree in the Department of Computer Graphics
and Multimedia, Faculty of Information Technology,
BUT.
His research focuses on computer vision, particularly
traffic analysis -- detection and re-identification of vehicles from surveillance cameras.
\end{IEEEbiography}
\vspace{-0.35cm}
\begin{IEEEbiography}[{\includegraphics[width=1in,height=1.25in,clip,keepaspectratio]{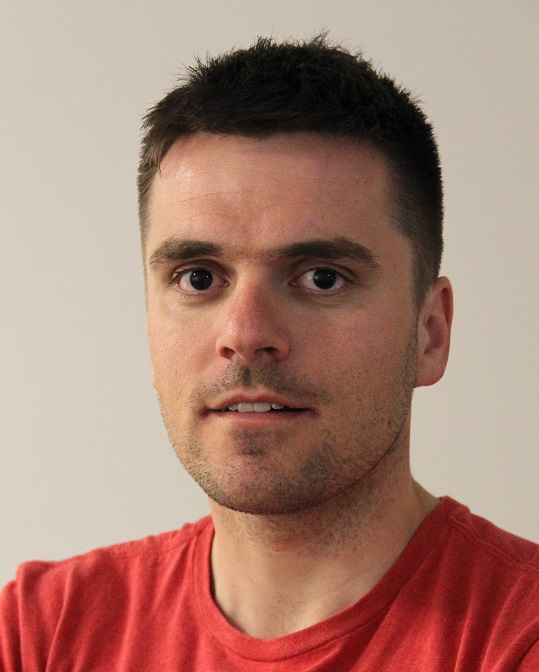}}]{Lukáš Maršík}
received his M.S. degree from
Brno University of Technology (BUT), Brno, Czech
Republic. He is currently working toward the Ph.D.
degree in the Department of Computer Graphics
and Multimedia, Faculty of Information Technology,
BUT.
His research interests include embedded signal processing, acceleration and radar signal processing with focus on intelligent traffic surveillance.
\end{IEEEbiography}
\vspace{-0.35cm}
\begin{IEEEbiography}[{\includegraphics[width=1in,height=1.25in,clip,keepaspectratio]{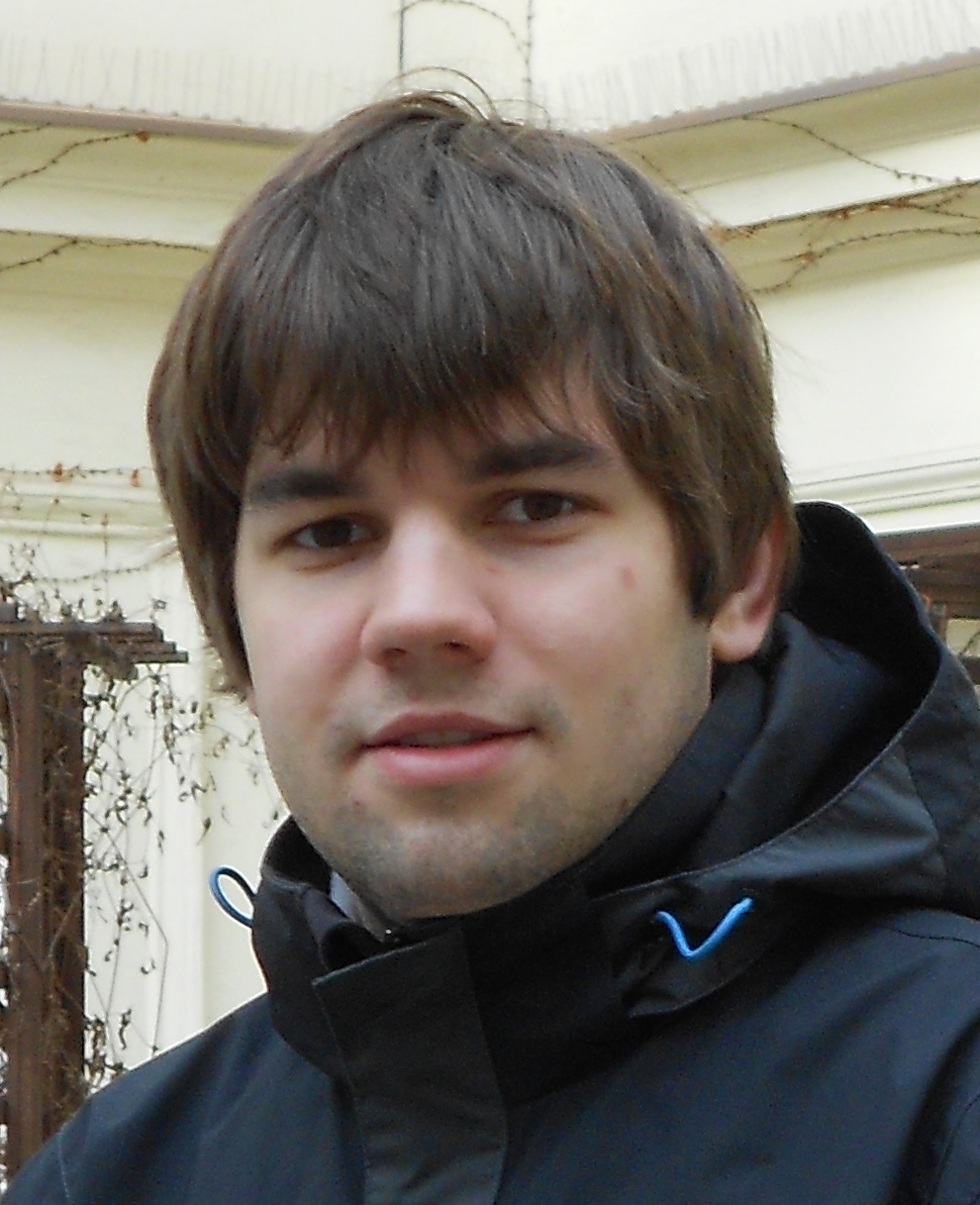}}]{Adam Široký}
	received his M.S. degree from Faculty of Information Technology,
	Brno University of Technology (BUT), Czech
	Republic. He is currently working toward the Ph.D.
	degree in the Department of Computer Graphics
	and Multimedia, Faculty of Information Technology,
	BUT.
	His research focuses on computer vision, particularly
	traffic analysis -- detection of vehicles from stereo camera.
\end{IEEEbiography}
\vspace{-0.35cm}
\begin{IEEEbiography}[{\includegraphics[width=1in,height=1.25in,clip,keepaspectratio]{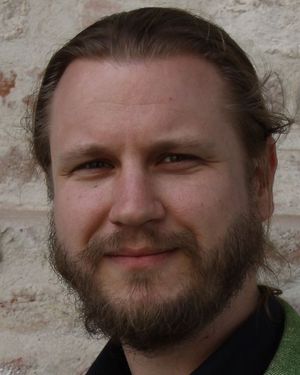}}]{Adam Herout}
	received his PhD from Faculty of Information Technology, Brno University of Technology, Czech Republic, where he works as a full professor and leads the Graph@FIT research group.  His research interests include fast algorithms and hardware acceleration in computer vision, with his focus on automatic traffic surveillance.
\end{IEEEbiography}
\vspace{-0.35cm}
\begin{IEEEbiography}[{\includegraphics[width=1in,height=1.25in,clip,keepaspectratio]{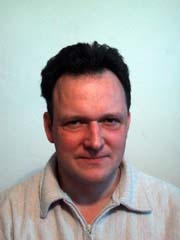}}]{Pavel Zemčík}
	received his PhD degree from the Faculty
	of Electrical Engineering and Computer Science, Brno
	University of Technology, Czech Republic. He works as a
	full professor, dean, and member of the Graph@FIT group
	at the Department of Computer Graphics and Multimedia at FIT, Brno University of Technology. His interests
	include computer vision and graphics algorithms,
	acceleration of algorithms, programmable hardware,
	and also applications.
\end{IEEEbiography}


\vfill


\end{document}